\documentclass{arxiv_template}

\PassOptionsToPackage{compress, round}{natbib}

\usepackage[utf8]{inputenc} % allow utf-8 input
\usepackage[T1]{fontenc}    % use 8-bit T1 fonts
\usepackage{hyperref}       % hyperlinks
\usepackage{url}            % simple URL typesetting
\usepackage{booktabs}       % professional-quality tables
\usepackage{amsfonts}       % blackboard math symbols
\usepackage{nicefrac}       % compact symbols for 1/2, etc.
\usepackage{microtype}      % microtypography
\usepackage{xcolor}         % colors

\definecolor{darkblue}{rgb}{0, 0, 0.5}
\hypersetup{breaklinks=true,colorlinks=true,citecolor=uscred}

\usepackage{graphicx}
\usepackage[dvipsnames]{xcolor}
\usepackage[most]{tcolorbox}
\definecolor{bboxbg}{HTML}{F1F4F7}
\usepackage[suppress]{color-edits}
\addauthor[Woody]{wg}{pink}
\addauthor[Julian]{j}{brown}
\addauthor[Deqing]{df}{Maroon} 
\addauthor[Oliver]{ol}{Purple} %northwestern? lol
\addauthor[Willie]{wn}{teal}
\addauthor[Robin]{rj}{green}
\addauthor[Vatsal]{vs}{red}

\usepackage{amsmath,amsthm,amssymb,amsfonts}
\usepackage{titletoc}

\renewcommand{\t}[1]{\texttt{#1}}

\usepackage{multirow}
\usepackage{algorithm}
\usepackage{algpseudocode}
\usepackage{subcaption}
\usepackage{wrapfig}

\definecolor{yelloworange}{RGB}{255,180,0}
\usepackage{amssymb,enumitem}
\usepackage{nicematrix}
\usepackage{makecell}
\usepackage{adjustbox}

\usepackage{xparse,xstring}   % colour + robust arg handling

\definecolor{Pos}{RGB}{0,158,115}    % bluish‑green   (good ↑)
\definecolor{Neg}{RGB}{213,94,0}     % vermilion      (bad  ↓)

\NewDocumentCommand{\score}{m m g}{%
  \ensuremath{%
    #1
    \IfNoValueTF{#3}{}{%
      \IfStrEq{#3}{+0.0}{%
        \textcolor{gray}{~~(#3)}%
      }{%
        \IfBeginWith{#3}{-}{%
          \textcolor{Neg}{~~(#3)}%
        }{%
          \textcolor{Pos}{~~(#3)}%
        }%
      }%
    }%
  }%
}

\usepackage{duckuments}
\DeclareMathOperator*{\argmax}{argmax}

\algnewcommand{\LeftComment}[1]{\Statex \(\triangleright\) #1}
\title{Textual Steering Vectors Can Improve Visual Understanding in Multimodal Large Language Models}

\newcommand{\aspace}{\hspace{2em}}
\newcommand{\equalcontrib}{{$^\star$}}

\author{\textbf{Woody Haosheng Gan}\equalcontrib\aspace 
\textbf{Deqing Fu}\equalcontrib\aspace 
\textbf{Julian Asilis}\equalcontrib\aspace 
\textbf{Ollie Liu}\equalcontrib\\
\textbf{Dani Yogatama} \aspace 
\textbf{Vatsal Sharan} \aspace 
\textbf{Robin Jia} \aspace 
\textbf{Willie Neiswanger}\vspace{2mm}\\
University of Southern California\\
\texttt{\{woodygan, deqingfu, asilis\}@usc.edu, me@ollieliu.com}\\
\texttt{\{yogatama, vsharan, robinjia, neiswang\}@usc.edu}
}
\contribution[{\star}]{Equal Contribution}

\abstract{Steering methods have emerged as effective and targeted tools for guiding large language models' (LLMs) behavior without modifying their parameters. Multimodal large language models (MLLMs), however, do not currently enjoy the same suite of techniques, due in part to their recency and architectural diversity. Inspired by this gap, we investigate whether MLLMs can be steered using vectors derived from their text-only LLM backbone, via sparse autoencoders (SAEs), mean shift, and linear probing. We find that text-derived steering consistently enhances multimodal accuracy across diverse MLLM architectures and visual tasks. In particular, mean shift boosts spatial relationship accuracy on CV-Bench by up to +7.3\% and counting accuracy by up to +3.3\%, outperforming prompting and exhibiting strong generalization to out-of-distribution datasets. These results highlight textual steering vectors as a powerful, efficient mechanism for enhancing grounding in MLLMs with minimal additional data collection and computational overhead.}

\begin{document}

\maketitle

\section{Introduction} \label{sec:intro}

% {
% \color{red}
% \begin{enumerate}
%     \item Say steering is useful for LLMs, and transfer methods developed in LLMs steering could solve some VLM problems (for example, counting and spatial relationship).
%     \item We explore different ways to do this. We also show that meanshift seems the best mechanism and it aligns with the conclusions from llm steering literature? 
%     \item We show the interesting/surprising finding that...[can we say one of our findings is interesting/surprising? If so lets point that out].
%     \item Can take advantage of existing infrastructure and machinery for textual steering.
% \end{enumerate}

% }

Steering large language models (LLMs) via their internal representations has emerged as a lightweight, interpretable paradigm for eliciting safe and controllable behavior~\citep[\textit{inter alia.}]{iti,turner2023steering,sharkey2025open}. This approach provides a targeted, interpretable way to guide model outputs without parameter updates. %\vscomment{I'm not sure if this is the pitch to make, if we do this then should actually compare to fine-tuning? Instead can talk about the other interpretability benefits provided by steering?} \olcomment{Good point. Changed the phrasing here!}
However, similar steering approaches have not yet gained prominence for \textit{multimodal large language models} (MLLMs). This is in part due to their relative recency, as well as the heterogeneity of their architectures compared to text-only LLMs. For example, many steering methods assume access to a dataset of contrast pairs~\citep{marks2023geometry} to construct steering vectors, which may not be readily available for multimodal inputs.

To address these limitations, we develop a multimodal steering paradigm that is agnostic to model architecture and does not require specialized multimodal data. Noting that most MLLMs are adapted from a pretrained LLM backbone, we design multimodal steering techniques by repurposing existing language steering methods originally developed for the base (text-modality-only) LLM. This approach leverages existing LLM steering methods that are developed and tested broadly in the text domain. It has the potential to obviate the need for modality-specific modifications. In doing so, we extend the benefits of lightweight and interpretable steering to the multimodal setting with minimal overhead, paving the way for more adaptable and broadly applicable control of MLLMs.

Sitting at the core of our method is the observation that internal representations from a text-only LLM backbone can be repurposed to steer its multimodal counterpart. Specifically, we extract interpretable steering vectors from the text-only LLM backbone using techniques based on Sparse Autoencoders (SAEs), Mean Shift, and Linear Probing, and apply these vectors to the hidden states of the MLLM to enhance multimodal reasoning capabilities.

We evaluate our approach on several open-weight MLLMs across a diverse suite of visual reasoning tasks, including spatial relations and object counting. Our method consistently outperforms prompting, demonstrating the practical utility of leveraging text-only steering techniques in the multimodal regime. Consistent with prior work~\citep{marks2023geometry,wu2025axbench}, we also observe that mean shift outperforms sparse autoencoders (SAEs) in key aspects of our implementation. Interestingly, although direct prompting is most effective for controlling \textit{text-only} LLMs' behavior \citep{wu2025axbench}, direct prompting barely helps in improving \textit{multimodal} LLMs' visual reasoning capabilities. Our contributions are summarized as follows:

\begin{itemize}[leftmargin=7mm,itemsep=1mm, topsep=0em,, parsep=0pt]
\item We introduce a plug-and-play multimodal steering paradigm that is compatible with existing internal-representation-based LLM steering techniques.
\item We identify a transfer effect: representations from the text-only LLM backbone remain effective for steering its multimodal counterpart, even after vision-language post-training.
\item We demonstrate consistent performance gains across multiple MLLMs and task categories.  Importantly, we also show that textual steering vectors could generalize to out-of-distribution test sets and demonstrate significant performance gains. 
%\vsmargincomment{i feel like the OOD results are maybe the strongest, let's mention them more explicitly in contributions?}
\end{itemize}

% We hypothesize that this transfer is enabled by the backbone’s preserved semantic structure through multimodal finetuning~\citep{lieberum2024gemmascope}, which enables a new form of text-derived guidance for vision inputs.

% \olcomment{TODO: briefly summarize the matching mechanism and how it’s integrated with the steering method.} 

\begin{figure}[t]
    \centering
    \includegraphics[width=\linewidth]{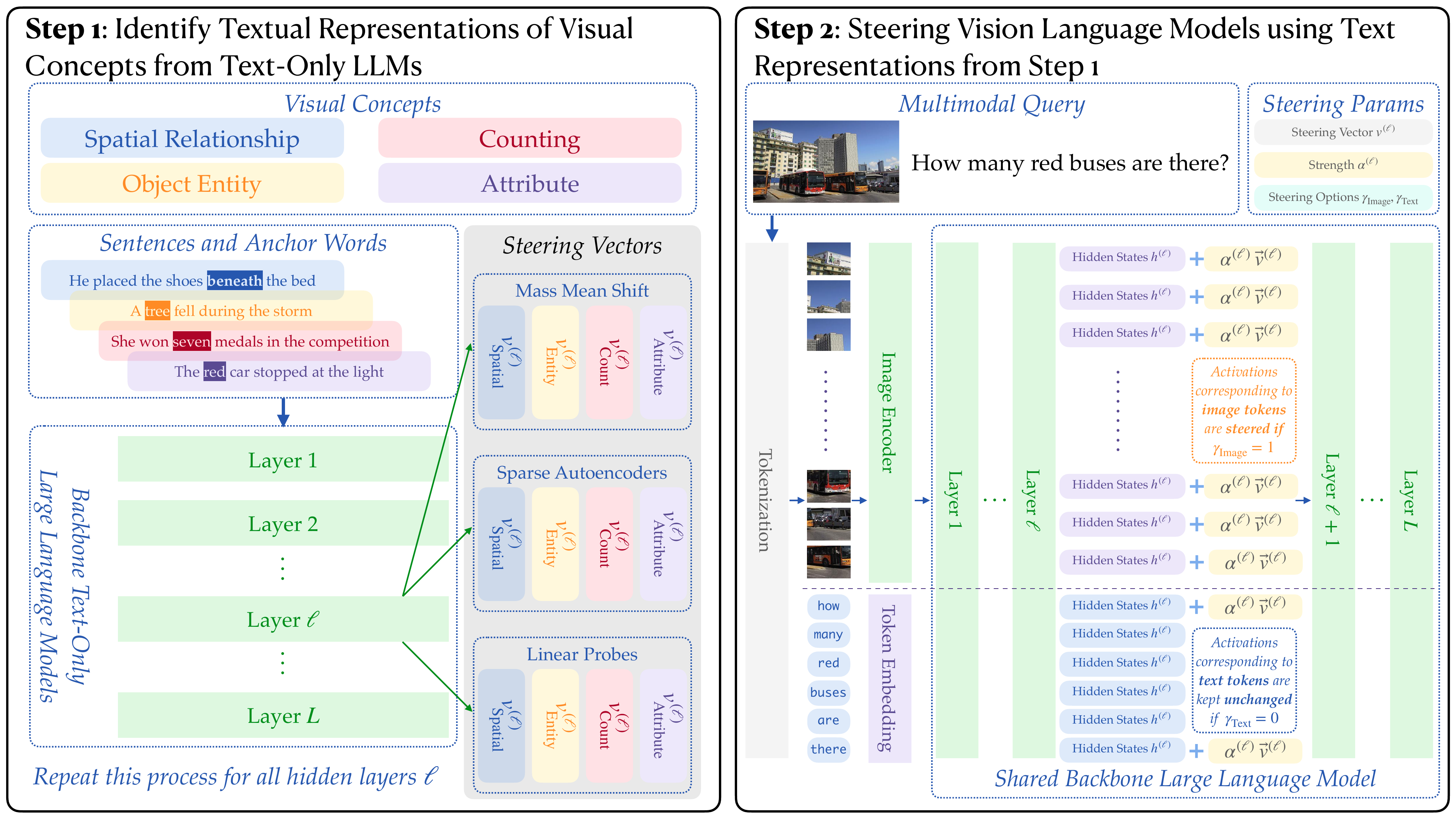}
    \caption{Overview of our steering methodology. For an MLLM with a text-only LLM backbone and a given image-bound prompt, we first identify the visual concept required to address the prompt (\emph{e.g}, spatial relationships, counting, object entity, or attributes). For each hidden layer $\ell$, we then determine corresponding steering vectors for the identified concept in the underlying LLM (\emph{e.g.} \texttt{Gemma2-2B} is the backbone LLM for the its multimodal counterpart \texttt{PaliGemma2-3B}). We study the selection of such vectors using mean shift, linear probing, and sparse autoencoders. Finally, we intervene these steering vectors on activations corresponding to the \emph{image tokens}, text tokens or both, depending on our choices of $\gamma_\mathrm{Image}$ and $\gamma_\mathrm{Text}$ from the original image-bound prompt. 
    %\dfcomment{overview of the pipeline. I think the right part is a bit repetitive. Need to emphasize the multimodal model shares the same base LLM as the text-only model. For the caption, I'll take another pass later.} 
    %\jcomment{Gave caption a shot, might be too long.}
    } 
    \label{fig:overview}
\end{figure}

\section{Related Works}\label{sec:related}

% We review representation-based steering techniques that underlie our text-driven approach, explore how MLLMs are built from unimodal backbones, and examine evidence for cross-modal representation transfer.

\textbf{Representation-Based Steering} methods are an effective family of methods for steering LLMs, often in two stages. \textbf{First}, they identify model components that influence target behaviors, using probing directions~\citep{iti,zou2023representation}, 
activation differences~\citep{iti,turner2023steering,panickssery2023steering,marks2023geometry,lee2024programming}, or lifted monosemantic features via SAEs~\citep{lieberum2024gemmascope,gao2025scaling,templeton2024scaling,marks2025sparse} and their variants~\citep{dunefsky2024transcoders}, among other techniques. \textbf{Second}, they adjust steering hyperparameters to balance desiderata such as truthfulness~\citep{lin-etal-2022-truthfulqa,hernandez2023inspecting,iti}, helpfulness~\citep{zou2023representation}, and quality. 

% \textbf{Activation Intervention} methods are an effective family of methods for steering LLMs toward desired behaviors. Most approaches follow a two-stage process. \textbf{First}, they identify model components that significantly contribute to target behaviors. This is typically done via probing directions~\citep{iti,zou2023representation}, steepest descent directions~\citep{subramani2022extracting,hernandez2023inspecting}, activation differences~\citep{iti,turner2023steering,panickssery2023steering,lee2024programming}, or lifted monosemantic features using methods such as sparse autoencoders (SAEs)~\citep{lieberum2024gemmascope,gao2025scaling,templeton2024scaling,marks2025sparse} and transcoders~\citep{dunefsky2024transcoders}. \textbf{Second}, they adjust steering hyperparameters to balance a mixture of desiderata, such as truthfulness~\citep{lin-etal-2022-truthfulqa,hernandez2023inspecting,iti}, helpfulness~\citep{zou2023representation}, sycophancy~\citep{turner2023steering}, and generation quality. These methods often rely on a training dataset possibly paired samples that reflect the intended behaviors.

While widely studied in LLMs, applying activation intervention to MLLMs remains elusive. To our knowledge, the only such effort is the VTI method of~\citet{liu2025reducing}, which extends LLM steering pipelines by constructing intervention vectors from paired multimodal inputs and applying them to both visual and textual representations. In contrast, we show that interventions vectors constructed solely from text inputs in the unimodal LLM can influence the MLLM’s multimodal behavior. This result highlights an underexplored form of cross-modal transfer enabled by the preserved semantics~\citep{lieberum2024gemmascope} of the text backbone.

\textbf{Shared Semantics} refer to the representations unifying heterogeneous modalities of the same content, as identified across languages in multilingual LLMs~\citep{artetxe2019cross,wendler2024llamas,wu2024semantic} and text/vision inputs in multimodal models~\citep{huh2024platonic,luo2024task,wu2024semantic}. Our work studies the transfer of steering effect across different modalities and training stages.
% While previous studies typically focus on abstract, task-specific representations, we leverage shared semantics at an {\em algorithmic} level, demonstrating that steering directions from the text-only backbone can influence MLLM behavior when applied solely to image tokens. 
% This suggests that semantic transfer can operate algorithmically, as hinted at by interpretability studies~\citep{wang2022interpretability,hanna2023does}.

% \textbf{Shared Semantics} denote shared representations across heterogeneous views of the same underlying data. Prior work has observed unified representations across languages in multilingual LLMs~\citep{artetxe2019cross,wendler2024llamas,wu2024semantic}, and similar findings extend to textual and visual modalities~\citep{huh2024platonic,wu2024semantic,luo2024task}. These studies primarily investigate abstract, task-level representations derived from in-context learning (ICL) examples or instructions. In contrast, we focus on activation interventions derived solely from the unimodal text-only backbone, and apply these directly to image tokens within the VLM. Unlike prior work, we show that behavioral steering directions learned in the language domain can modulate multimodal behavior—even when applied exclusively to the visual input. Our findings hint at the possibility of transferring semantic representations at an \textit{algorithmic} level~\citep{wang2022interpretability,hanna2023does}.

\textbf{Multimodal Large Language Models} are commonly developed by endowing a backbone LLM with visual processing components and fine-tuning on multimodal datasets, with some exceptions still pretrained from scratch \citep{chameleon,olmo20242,chen2025janus}. Using an LLM backbone typically involves projecting the outputs of an image encoder (\emph{e.g.}, \citet{dosovitskiy2020image,zhai2023sigmoid}) to the same dimension as the underlying LLM by an MLP, and concatenating the resulting image/text tokens as input to the LLM. The model can then be finetuned on multimodal data, possibly with frozen layers (\emph{e.g.}, in the LLM) to preserve pretrained knowledge.

\section{Toy Example} \label{sec:toy}

\begin{wrapfigure}{r}{0.4\textwidth}
\vspace{-5mm}
  \begin{center}
    \includegraphics[width=\linewidth]{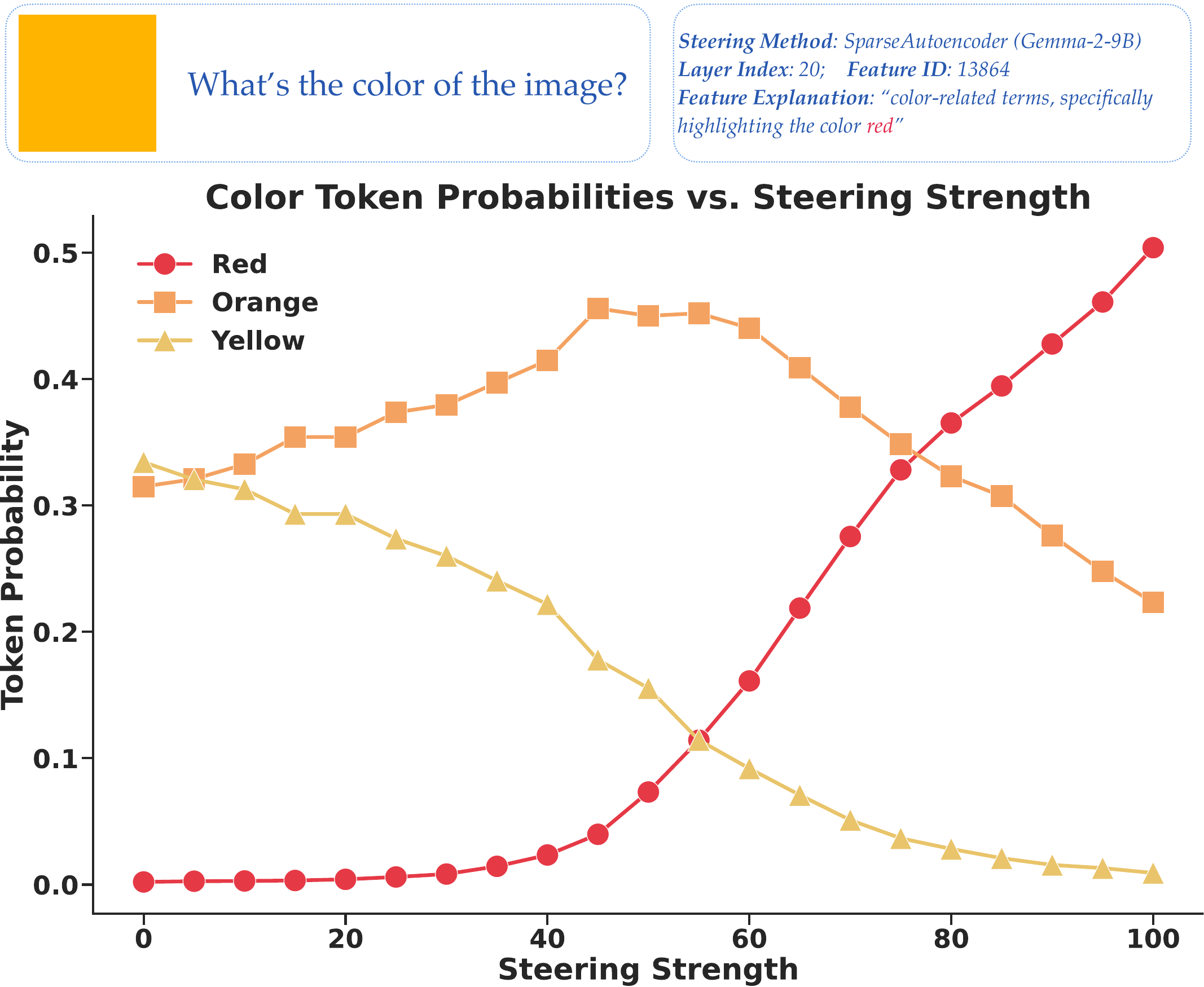}
  \end{center}
  \caption{Effect of steering strength on color token probabilities.}
  \label{fig:color_intervention} 
  \vspace{-2mm}
\end{wrapfigure}

To demonstrate that textual representations can effectively intervene in visual understanding, we conduct a simple color perception experiment using GemmaScope \citep{lieberum2024gemma} for \texttt{Gemma-2-9B} for feature extraction and \texttt{PaliGemma2-10B-mix-448} \citep{beyer2024paligemma} as our target model.
We present the model with a yellow-orange image (whose RGB hex code is {\color{yelloworange}\#FFB400}) and manipulate its perception by intervening in the hidden representations. Specifically, we find the normalized {\color{red} red} vector from GemmaScope and we add this vector to the hidden states of image tokens at layer 20 as follows: $h'_\mathrm{image} = h_\mathrm{image} + \alpha \cdot v_\mathrm{red}$, where $\alpha$ is the scale factor controlling intervention strength. Figure~\ref{fig:color_intervention} shows how increasing the scale factor shifts perception along a color spectrum: initially yellow-orange dominates, then orange peaks at scale factor 50, and finally red becomes dominant beyond scale factor 75. This demonstrates that textual features can integrate with and modify visual understanding, supporting our hypothesis of unified cross-modal representations within these models. We include more color examples in Appendix \ref{appendix:color_examples}. 
\section{Methods} \label{sec:methods}

\begin{figure}[t]
\centering
\includegraphics[width=0.8\linewidth]{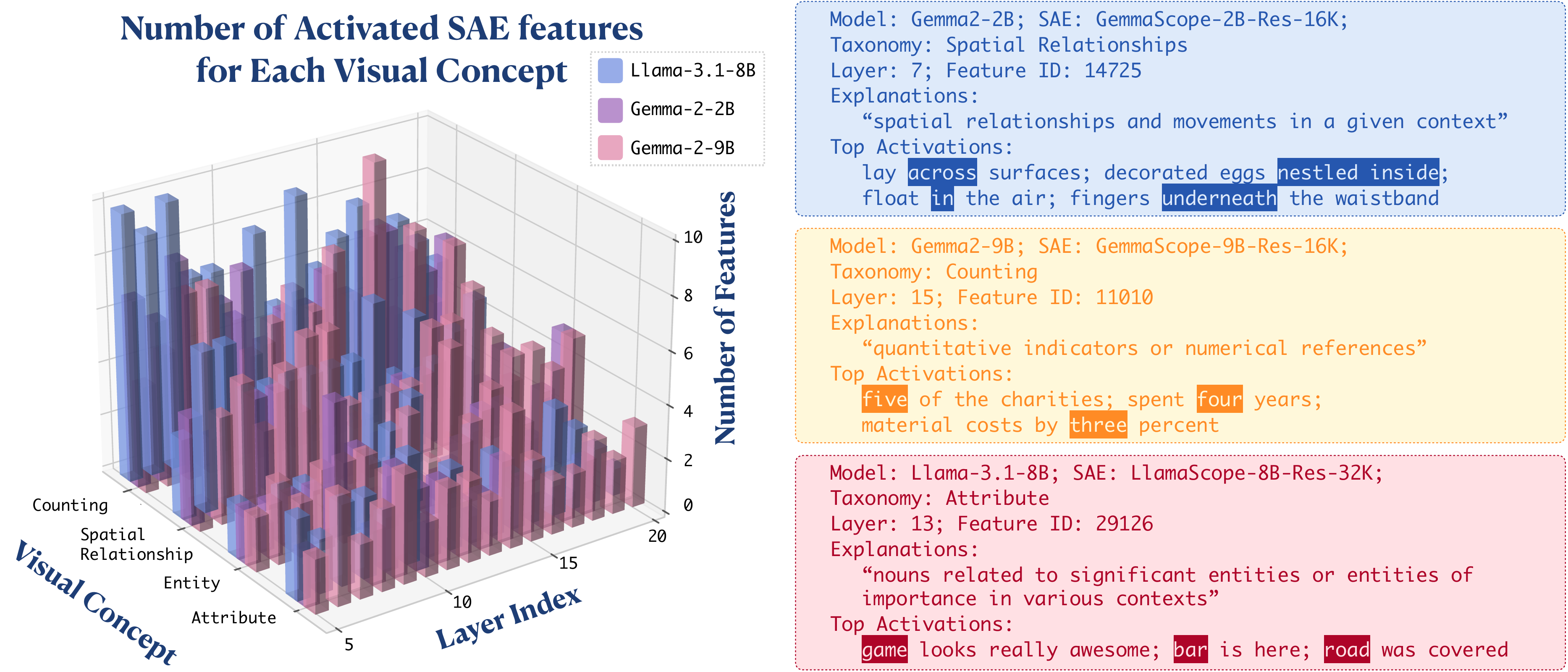}
\caption{\textbf{Left}: Number of SAE features associated with each taxonomy (counting, spatial relationship, entity, and attribute) across the layers of Llama-3.1-8B, Gemma2-2B, and Gemma2-9B. Notably, SAE features for such visual concepts are sparse, numbering fewer than $10$ across 16k total SAE features (Gemma2-2B/9B) or 32k features (Llama-3.1-8B).
%\vscomment{increase font size for figure?}\dfcomment{We lost this server...Will do some post editing.}
\textbf{Right}: Examples of features corresponding to visual concepts, identified by the layer whose activation space they inhabit and their (arbitrary) feature ID. The feature's explanation summarizes its semantic meaning, as evidenced by the tokens and contexts on which it attains the greatest activations.
}
\label{fig:sae-features}
\end{figure}

Building on our successful intervention in Section~\ref{sec:toy}, where we demonstrate that textual representations can effectively steer visual understanding, we now explore more systematic approaches to improve MLLMs' visual reasoning. Despite their growing success, Multimodal Large Language Models (MLLMs) still stumble on seemingly simple visual queries—miscounting objects, confusing left with right, and mishandling compositional prompts \citep{fu2024blink}. When the same problems are posed in pure text, foundation models are far better than when they are asked to reason over images \citep{fu2024isobench}.

A promising remedy in the text‑only world is the use of steering vectors: compact directions in a model’s activation space that are learned to emphasize or suppress specific behaviors. At inference time, the steering vectors are simply added to one or more hidden layers, where the best layer and intervention strength could be found via grid search. 

This observation motivates the central question of this section: 
\begin{center}
    \textit{\textbf{Can steering mechanisms for textual representations rectify the shortcomings of MLLMs?}}
\end{center}

We now review several techniques for extracting textual steering vectors from LLMs: sparse autencoders (SAE), mean shift, and linear probing (Probe). These techniques constitute Step 1 of our steering methodology, as displayed in Figure~\ref{fig:overview}. In Section~\ref{ssec:prompting} we briefly discuss prompting, which will serve as a baseline for the three steering methods.
The transferability of these methods to MLLMs (\emph{i.e.}, Step 2) will be the subject of Section~\ref{sec:steering_improves}.

\subsection{Crafting Datasets to Extract Steering Vectors}
\label{ssec:steering_dataset}
In order to find high-level textual representations for visual concepts, we first identify four important taxonomies for static images, as considered by prior work \citep{huang2023t2i,lin2024evaluating,fu2025tldr}: spatial relationship, counting, attribute, and entity.  For each visual concept, we curate \textit{a small set of sentence-anchor pairs}, where each pair contains a sentence exhibiting the visual concept and the specific anchor word representing that concept. Examples are shown in Table~\ref{tab:taxonomy_sample} of Appendix~\ref{Appendix:SAEs}. These sentence-anchor pairs serve as the foundation for all three steering vector extraction methods described in the following subsections.
\subsection{Sparse Autoencoders (SAE)}
\label{ssec:sae}
Sparse autoencoders reconstruct the activations of an LLM's hidden layer using an MLP with a single hidden layer and a sparsity penalty on the hidden layer. More precisely, let $x = h^{(\ell)}(t) \in \mathbb{R}^D$ be the model activations for a token $t$ at layer $\ell$ in an LLM. A SAE reconstructs $x$ as $\hat{x} = b_{\text{dec}} + \sum_{i=1}^{F} f_i(x) \, W^{\text{dec}}_{\cdot,i}$, where \( b_{\text{dec}} \in \mathbb{R}^D \) and \( W^{\text{dec}} \in \mathbb{R}^{D \times F} \) are learned decoder weights, and \( f_i(x) \) is the activation corresponding to feature \( i \). Feature activations are computed using learned encoder weights $W^{\text{enc}} \in \mathbb{R}^{F \times D}$ and $b^{\text{enc}} \in \mathbb{R}^F$ as $f_i(x) = \sigma\Big( W^{\text{enc}}_{i,\cdot}\, x + b^{\text{enc}}_i \Big)$, where $\sigma$ denotes an activation function of choice, \emph{e.g.}, ReLU or JumpReLU. 

The model is trained by minimizing the loss function $L = \mathbb{E}_{x}\left[ \|x - \hat{x}\|_2^2 + \lambda \sum_{i=1}^{F} f_i(x) \, \|W^{\text{dec}}_{\cdot,i}\|_2 \right]$, \emph{i.e.}, $L_2$-reconstruction error and $L_1$-regularization on feature activations.
% ,  weighted by the norm of the corresponding decoder weight vector and controlled by the regularization parameter \( \lambda \)
In this formulation, unit-normalized decoder weight vectors $v^{(\ell)}_{i} := \frac{W^{\text{dec}}_{\cdot,i}}{\|W^{\text{dec}}_{\cdot,i}\|_2}$ serve as feature directions and $\alpha^{(\ell)}_i(t) := f_i(h^{(\ell)}(t)) \, \|W^{\text{dec}}_{\cdot,i}\|_2$ as the activation strength of $v^{(\ell)}_i$ on token $t$.

For our experiments, we leverage pretrained SAEs—GemmaScope \citep{lieberum2024gemma} for Gemma-2-2B and Gemma-2-9B, and LlamaScope \citep{he2024llamascope} for Llama-3.1-8B. 
We emphasize that \textit{training SAEs is very costly} and that an advantage of steering using text-based vectors is that one can leverage existing interpretability tools.
%Using our sentence-anchor pairs, we identify features with high activations on anchor words. 

We then use the sentence-anchor pairs described in Section~\ref{ssec:steering_dataset} to identify feature directions corresponding to the ideal visual concepts using Algorithm \ref{alg:sae-find-feature}. We employ a two-stage procedure which, at the first stage, finds the top $n$ activated features for anchor words $w_j$ in sentences $s_j$. Interestingly, as shown in Figure~\ref{fig:sae-features}, we find that each visual concept activates only a limited number of SAE features, indicating a sparse encoding of these concepts. 
At the second stage, We then verify their relevance to the target visual concepts. We use \texttt{o3-mini} \citep{openai_o3mini} to verify that these features indeed align with the desired visual concept $\mathcal C$.  When we prompt \texttt{o3-mini} for verification, we craft prompts to include both the explanation for the candidate feature vector $v^{(\ell)}_i$, and sample top activated tokens (see Figure~\ref{box:o3_prompt} for the prompting template). We find that \texttt{o3-mini} can indeed filter out features unrelated to the desired visual concepts. Finally, we average these relevant feature vectors to create a single steering vector for each visual concept at each layer. Additional details are provided in Appendix~\ref{Appendix:SAEs}.

\begin{algorithm}[t]
\caption{Find Textual Representations for Visual Concepts using SAEs}\label{alg:sae-find-feature}
\begin{algorithmic}
\Require Desired visual concepts $\mathcal C$. Layer index $\ell$. 
\Require Sentence and anchor word pairs $\{(s_1, w_1), \cdots, (s_K, w_K)\}$. 
\Require Pretrained SAEs at layer $\ell$.
\LeftComment{Find top activations and their corresponding SAE feature vectors.}
\State $\mathcal V_0 = \{\}$ 
\For{each $(s_j, w_j)$}
    \State $\{\alpha^{(\ell)}_i(w_j), v^{(\ell)}_i\} \leftarrow$ Pass $s_j$ into the pretrained SAE 
    \State $\{v^{(\ell)}_{i_1}, \cdots v^{(\ell)}_{i_n}\} \leftarrow \mathrm{Top}_n \{\alpha^{(\ell)}_i(w_j), v^{(\ell)}_i\}$ ranked by activation strength $\alpha^{(\ell)}_i(w_j)$
    \State $\mathcal V_0 \leftarrow \mathcal V_0 \cup \{v^{(\ell)}_{i_1}, \cdots v^{(\ell)}_{i_n}\}$
\EndFor

\LeftComment{Filter out noisy SAE feature vectors.}
\State $\mathcal V = \{\}$
\For{each $v^{(\ell)}_i \in \mathcal V_0$}
    \State Find the explanation $e$ and top activated tokens $\{t_1, \cdots, t_p\}$ for $v^{(\ell)}_i$
    \If{\texttt{o3-mini}(VerificationPrompt, $e$, $\{t_1, \cdots, t_p\}$, $\mathcal C$) is True}
    \State $\mathcal V \leftarrow \mathcal V \cup \{v^{(\ell)}_i\}$
    \EndIf
\EndFor

\LeftComment{Aggregate SAE vectors to one steering vector.}
\State $v^{(\ell)} = \frac{1}{|\mathcal{V}|}\sum_{u \in \mathcal V} u$
\State \Return $v^{(\ell)}$
\end{algorithmic}
\end{algorithm}

\subsection{Mean Shift} \label{ssec:meanshift}
An alternative approach to identify feature directions for visual concepts is through mean shift analysis, which has shown surprising effectiveness for steering in LLMs \citep{marks2023geometry, wu2025axbench}.
Given a text-only large language model $h$, for each taxonomy $\mathcal{T} \in \{\mathrm{spatial}$ $\mathrm{relationship}, \mathrm{counting}, \mathrm{attribute}, \mathrm{entity}\}$ and layer $\ell$, we use the same set of sentence-anchor pairs $\{(s_1, w_1), \ldots, (s_K, w_K)\}$ as described in~\ref{ssec:steering_dataset}. We compute the mean shift vector between hidden states of the residual stream as:
\begin{align*}
m^{(\ell)}_{\mathcal{T}} = \frac{1}{K}\sum_{j=1}^{K} h^{(\ell)}(w_j) - \frac{1}{|\mathcal{S}_{\neg\mathcal{T}}|}\sum_{t \in \mathcal{S}_{\neg\mathcal{T}}} h^{(\ell)}(t)\, ,
\end{align*}
where $h^{(\ell)}(w_j)$ represents the residual stream activation of the anchor word $w_j$ at layer $\ell$ and $\mathcal{S}_{\neg\mathcal{T}}$ is a control set of non-anchor tokens from the same sentences containing the anchor words. By using tokens from the same contextual environment but excluding the anchor words themselves, we isolate the specific representation of the visual concept from the general contextual information. We deliberately refrain from normalizing the vector $m^{(\ell)}_{\mathcal{T}}$, preserving its magnitude relative to the original hidden states.
% (currently it's just all tokens except special tokens and the anchor word token)

\begin{figure}[t]
    \centering
    \includegraphics[width=0.4\linewidth]{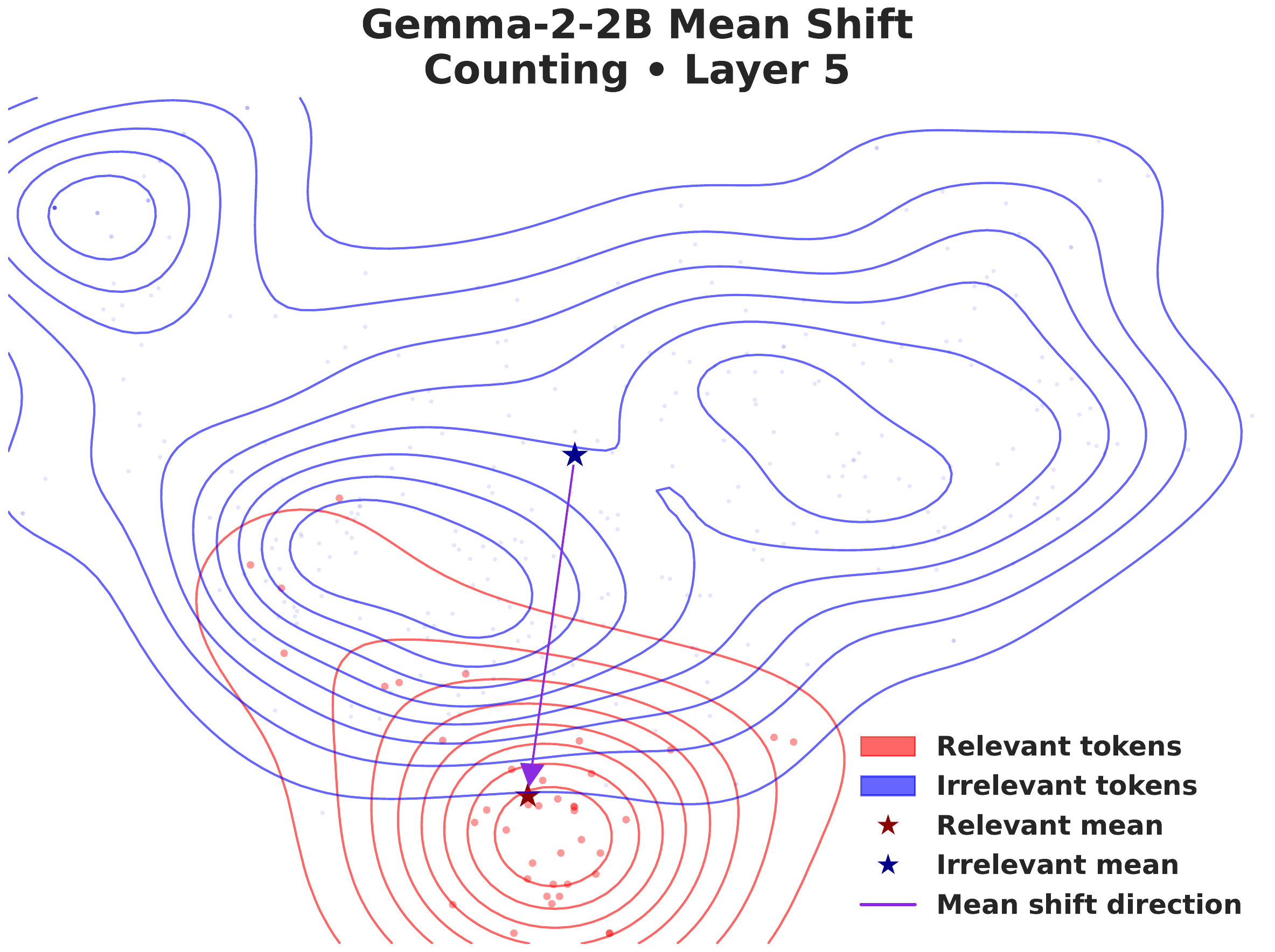} \hspace{5mm}
    \includegraphics[width=0.4\linewidth]{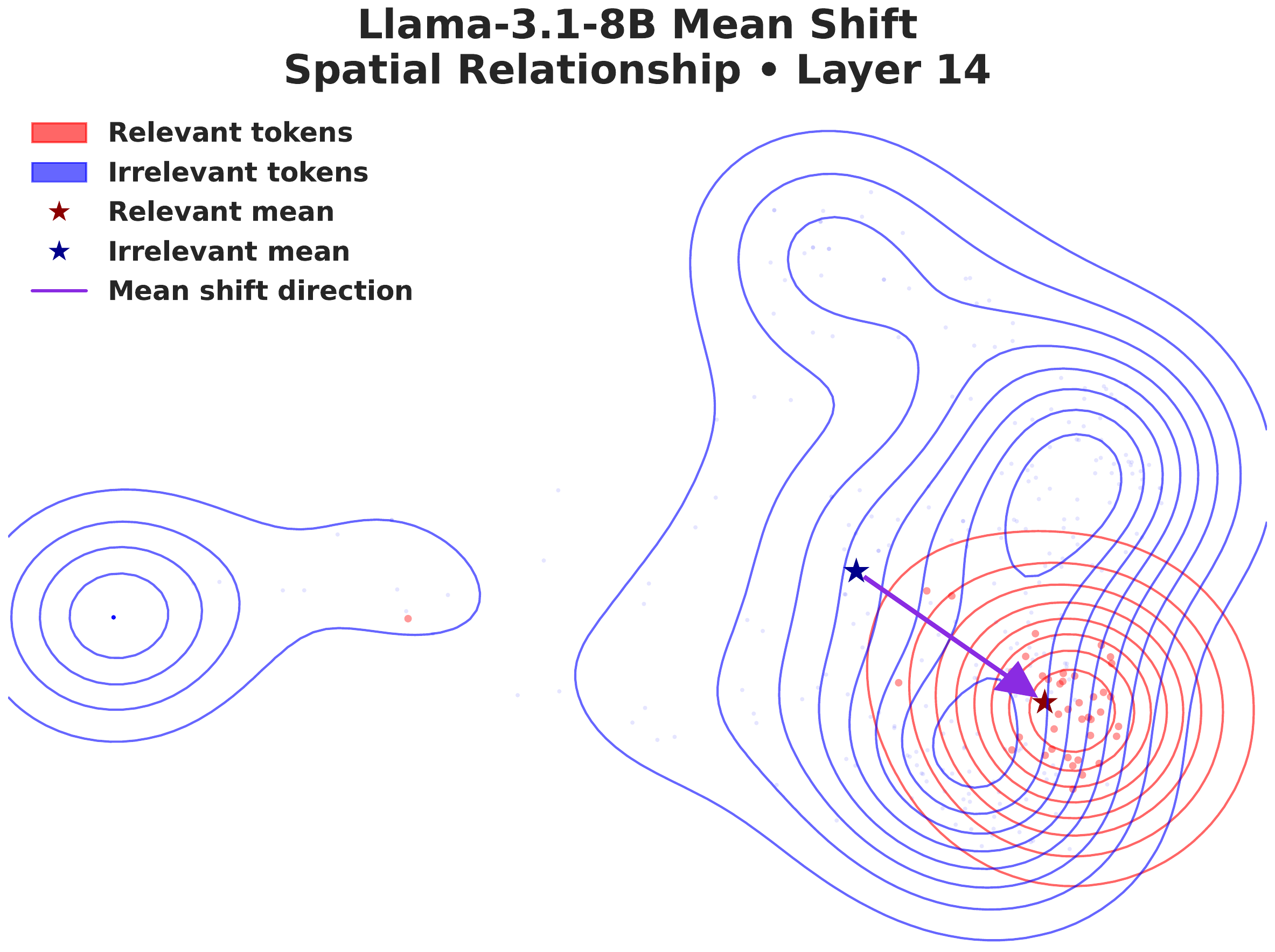} 
    \caption{\textbf{Left:} Depiction of mean shift method for the counting feature for \texttt{Gemma-2-2B}. The mean shift direction points from the mean hidden state of irrelevant tokens to the mean hidden state of relevant tokens (\emph{i.e.}, counting-related tokens). Activations are projected to two dimensions for the sake of visualization. \textbf{Right:} Similar depiction for the spatial relationship feature of \texttt{Llama-3.1-8B}.}
    \label{fig:meanshift_vis}
\end{figure}

\subsection{Linear Probing}
The third technique we employ is linear probing (a.k.a probe), training a classifier from the $\ell$-th layer of a model \citep{alain2016understanding, park2024linear}. Let $h^{(\ell)}(t) \in \mathbb{R}^D$ denote the activations for token $t$ at layer $\ell$. Using the same sentence-anchor pairs described in Section~\ref{ssec:meanshift}, we train a linear function to distinguish between anchor word activations and a control set of non-anchor tokens from the same sentences, isolating representations specific to the taxonomy.

As the hidden state dimensionality often exceeds our sample size ($K < D$), we first project to dimension $d < K$ using PCA. With $Q \in \mathbb{R}^{d \times D}$ as the PCA matrix, the probe separates: 
\[ \{h^{(\ell)} (w_j) Q^\top\}_{j \leq K}, \qquad \text{ and } \qquad \{h^{(\ell)} (t) Q^\top\}_{t \in \mathcal{S}_{\neg\mathcal{T}}}, \] 
where $\{(s_1, w_1), \ldots, (s_K, w_K)\}$ are the sentence-anchor pairs for concept $\mathcal{T}$ and $\mathcal{S}_{\neg\mathcal{T}}$ is our control set. The learned normal vector $v \in \mathbb{R}^d$ (pointing toward taxonomy-relevant points) yields the final steering vector $v' = Q^{\top} v$. We use $d = K / 2$ in practice.
% \jcomment{Missing anything, @Woody?} \wgcomment{It looks good to me@}

\subsection{Prompting}
\label{ssec:prompting}

We now briefly discuss prompting, which, despite not employing a steering \emph{vector}, has displayed impressive abilities in text-only domains \citep{wu2025axbench}. In Section~\ref{sec:steering_improves}, prompting will serve as a baseline for comparison with the previous vector-based methods, which will furthermore shed light on whether the efficacy of prompting transfers to multimodal settings. 

For a given taxonomy $\mathcal{T}$, we generate a prompt meant to enhance an MLLM's visual reasoning ability with respect to $\mathcal{T}$ as follows: We first curate a collection of 30 prompts $\{p_1, \ldots, p_{30}\}$ of varying lengths by instructing \texttt{GPT-4o-2024-08-06} to generate instructions that will elicit improved reasoning with respect to $\mathcal{T}$, similar to the LLM-based prompt generation approach used in AxBench \citep{wu2025axbench}. We then do a grid search for all prompts on a training set and retain only the best-performing prompt $p_{\mathrm{best}}$ for use on the test set. The performance of $p_{\mathrm{best}}$ is thus treated as the performance of prompting. Refer to Appendix~\ref{Appendix:prompting} for further detail. 
\section{Steering Improves Multimodal LLMs}\label{sec:steering_improves}

\begin{table}[t]
  \centering
  \begin{adjustbox}{width=\textwidth}
  \NiceMatrixOptions{cell-space-top-limit=1pt,cell-space-bottom-limit=1pt} % Reduced vertical padding
  \setlength{\tabcolsep}{2pt} %
  \begin{NiceTabular}{l|cc|lll|lll}[
        code-before = {\rowcolors{3}{gray!12}{white}}
    ]
    \toprule
      \multirow{2}{*}{\centering\textsc{\textbf{Model}}} &
      \multicolumn{2}{c}{\textsc{\textbf{\makecell{Intervention\\Tokens}}}} &
      \multicolumn{3}{c}{\textsc{\textbf{Relation}}} &
      \multicolumn{3}{c}{\textsc{\textbf{Count}}} \\
      & \textsc{Text} & \textsc{Image} &
        \textsc{SAE} & \textsc{Probe} & \textsc{MeanShift} &
        \textsc{SAE} & \textsc{Probe} & \textsc{MeanShift} \\
    \midrule
    \multirow{5}{*}{\texttt{PaliGemma2-3B}}
      & \multicolumn{2}{c}{---} & \multicolumn{3}{c|}{\score{76.0}{3.49}} & \multicolumn{3}{c}{\score{59.3}{4.01}} \\
      & \checkmark &            & \score{{82.0}}{3.14}{+6.0} & \score{77.3}{3.42}{+1.3} & \score{{83.3}}{3.04}{+7.3} & \score{60.0}{3.99}{+0.7} & \score{62.0}{3.96}{+2.7} & \score{60.0}{3.99}{+0.7} \\
      &            & \checkmark & \score{{78.7}}{3.34}{+2.7} & \score{76.7}{3.44}{+0.7} & \score{{78.7}}{3.34}{+2.7} & \score{62.0}{3.96}{+2.7} & \score{60.7}{3.99}{+1.3} & \score{62.0}{3.96}{+2.7} \\
      & \checkmark & \checkmark & \score{{81.3}}{3.19}{+5.3} & \score{{78.7}}{3.34}{+2.7} & \score{{81.3}}{3.19}{+5.3} & \score{62.7}{3.95}{+3.3} & \score{62.0}{3.96}{+2.7} & \score{62.0}{3.96}{+2.7} \\
      & \multicolumn{2}{c}{\texttt{Prompting}} & \multicolumn{3}{c|}{\score{74.0}{}{-2.0}} & \multicolumn{3}{c}{\score{60.7}{}{+1.3}} \\ 
    \midrule
    \multirow{5}{*}{\texttt{PaliGemma2-10B}}
      & \multicolumn{2}{c}{---} & \multicolumn{3}{c|}{\score{79.3}{3.31}} & \multicolumn{3}{c}{\score{63.3}{3.98}} \\
      & \checkmark &            & \score{78.7}{3.34}{-0.7} & \score{77.3}{3.42}{-2.0} & \score{83.3}{3.19}{+4.0} & \score{63.3}{3.94}{+0.0} & \score{62.7}{3.95}{-0.7} & \score{64.0}{3.96}{+0.7} \\
      &            & \checkmark & \score{{79.3}}{3.31}{+0.0} & \score{{79.3}}{3.31}{+0.0} & \score{78.7}{3.34}{-0.7} & \score{63.3}{3.94}{+0.0} & \score{63.3}{3.94}{+0.0} & \score{64.7}{3.92}{+1.3} \\
      & \checkmark & \checkmark & \score{78.7}{3.34}{-0.7} & \score{78.0}{3.37}{-1.3} & \score{{83.3}}{3.09}{+4.0} & \score{64.0}{3.92}{+0.7} & \score{63.3}{3.94}{+0.0} & \score{63.3}{3.95}{+0.0} \\
      & \multicolumn{2}{c}{\texttt{Prompting}} & \multicolumn{3}{c|}{\score{77.3}{}{-2.0}} & \multicolumn{3}{c}{\score{62.7}{}{-0.7}} \\ 
    \midrule
    \multirow{5}{*}{\texttt{Idefics3-8B-Llama3}}
      & \multicolumn{2}{c}{---} & \multicolumn{3}{c|}{\score{73.3}{3.61}} & \multicolumn{3}{c}{\score{59.3}{4.01}} \\
      & \checkmark &            & \score{76.0}{3.49}{+2.7} & \score{78.0}{3.37}{+4.7} & \score{80.0}{3.27}{+6.7} & \score{58.7}{4.02}{-0.7} & \score{58.0}{4.03}{-1.3} & \score{60.0}{3.99}{+0.7} \\
      &            & \checkmark & \score{{78.0}}{3.37}{+4.7} & \score{72.7}{3.64}{-0.7} & \score{76.7}{3.45}{+3.3} & \score{60.0}{3.99}{+0.7} & \score{59.3}{4.01}{+0.0} & \score{60.7}{3.99}{+1.3} \\
      & \checkmark & \checkmark & \score{77.3}{3.42}{+4.0} & \score{{78.7}}{3.34}{+5.3} & \score{{80.7}}{3.24}{+7.3} & \score{62.0}{3.96}{+2.7} & \score{60.0}{3.99}{+0.7} & \score{60.7}{3.99}{+1.3} \\
      & \multicolumn{2}{c}{\texttt{Prompting}} & \multicolumn{3}{c|}{\score{75.3}{}{+2.0}} & \multicolumn{3}{c}{\score{59.3}{}{+0.0}} \\ 
    \bottomrule
  \end{NiceTabular}
  \end{adjustbox}
  \vspace{1mm}
  \caption{\textbf{Textual Steering Vectors Improve Multimodal LLMs' Visual Understanding}. Task-specific textual steering vectors reliably improve both spatial relation and counting performance across multimodal models. Combined interventions on both image and text tokens under the MeanShift framework yields the strongest gains.}
  \label{tab:combined_results}
\end{table}

Having established in \S \ref{sec:toy} that textual steering vectors applied to non-output tokens can alter the behavior of MLLMs, we now investigate whether the textual steering vectors we identified in \S \ref{sec:methods} can \textit{improve} visual understanding in MLLMs when applied to intermediate representations.

\subsection{Setup} \label{ssec:expt_setup}

\textbf{Models. } We mainly investigate PaliGemma2 models with 3B and 10B parameters (specifically, \texttt{PaliGemma2-3B-mix-448} and \texttt{PaliGemma2-10B-mix-448}, which we refer to as \texttt{PaliGemma2-3B} and \texttt{PaliGemma2-10B} for brevity), and the Idefics3 model with 8B parameters (\texttt{Idefic3-8B-Llama3}). These models were chosen because they are slightly different in architecture -- PaliGemma2 adopts the same prefix-LM masking strategy as PaliGemma, where the image tokens and textual instructions are cross-attended, whereas Idefics3 is an entirely autoregressive model following LLaVA. The steering vectors are extracted from \texttt{Gemma2-2B}, \texttt{Gemma2-9B}, and \texttt{Llama-3.1-8B} text-only LLMs, respectively, which serve as the pretrained backbones for the multimodal counterparts.

\textbf{Dataset. } In this section, we work with CV-Bench \citep{tong2024cvbench}, an evaluation dataset with 4 major sub-categories: \texttt{Count}, \texttt{Relation}, \texttt{Distance}, and \texttt{Depth}. CV-Bench contains a total of 2,638 data points with each sub-category containing around 700 samples. For each sub-category, we split the samples in to 500-600 training samples reserved for grid search, and 150 samples used for testing. 

\begin{figure}
    \centering
    \begin{subfigure}{0.4\textwidth}
    \includegraphics[width=\linewidth]{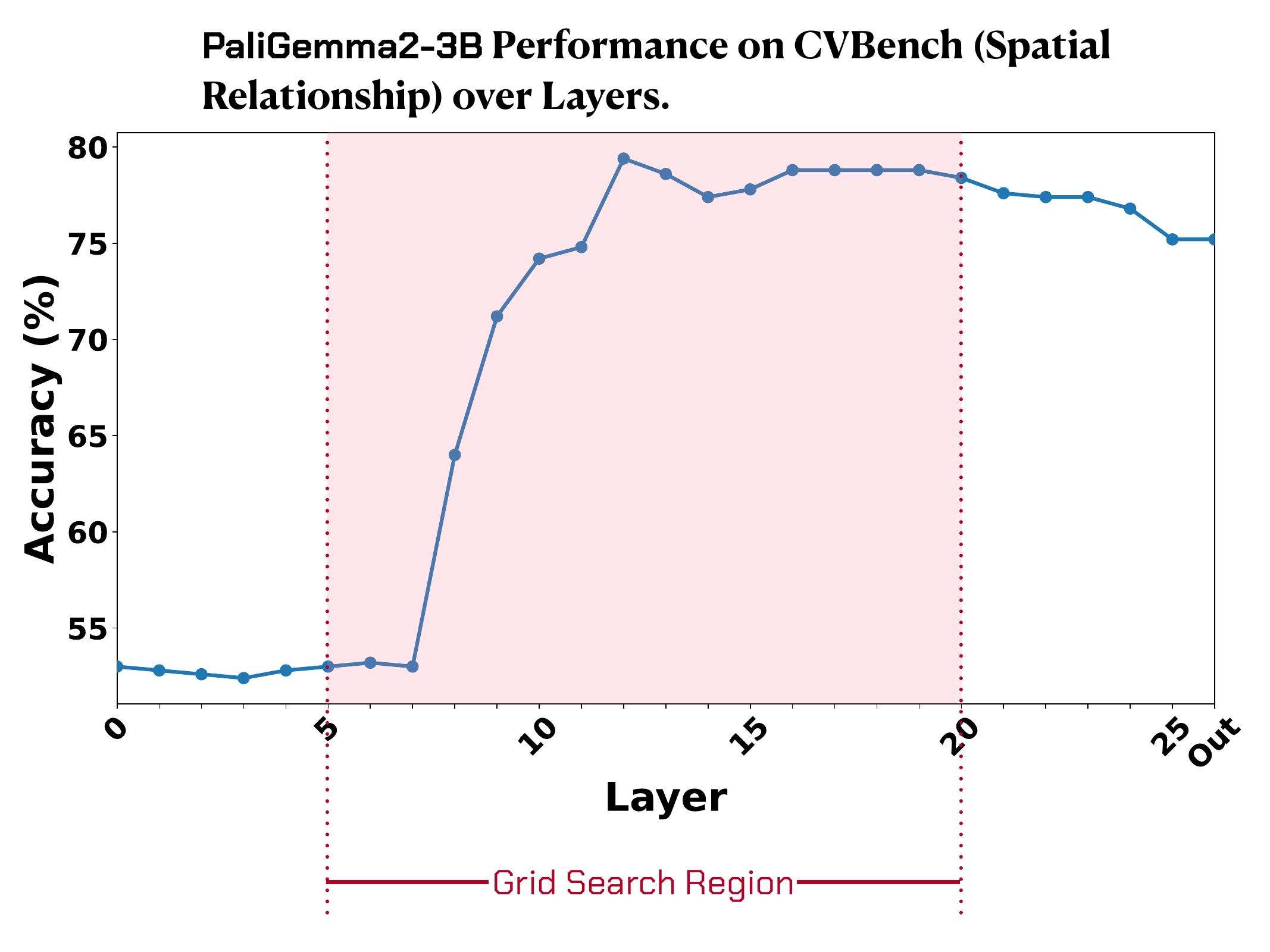}
        \caption{We zero out models' attention to image tokens after layer $\ell$ and measure model performance. This reveals when visual information is processed and allows efficient grid search.}
        \label{subfig:efficient_grid_search}
    \end{subfigure} \hspace{5mm}
    \begin{subfigure}{0.4\textwidth}
        \includegraphics[width=\linewidth]{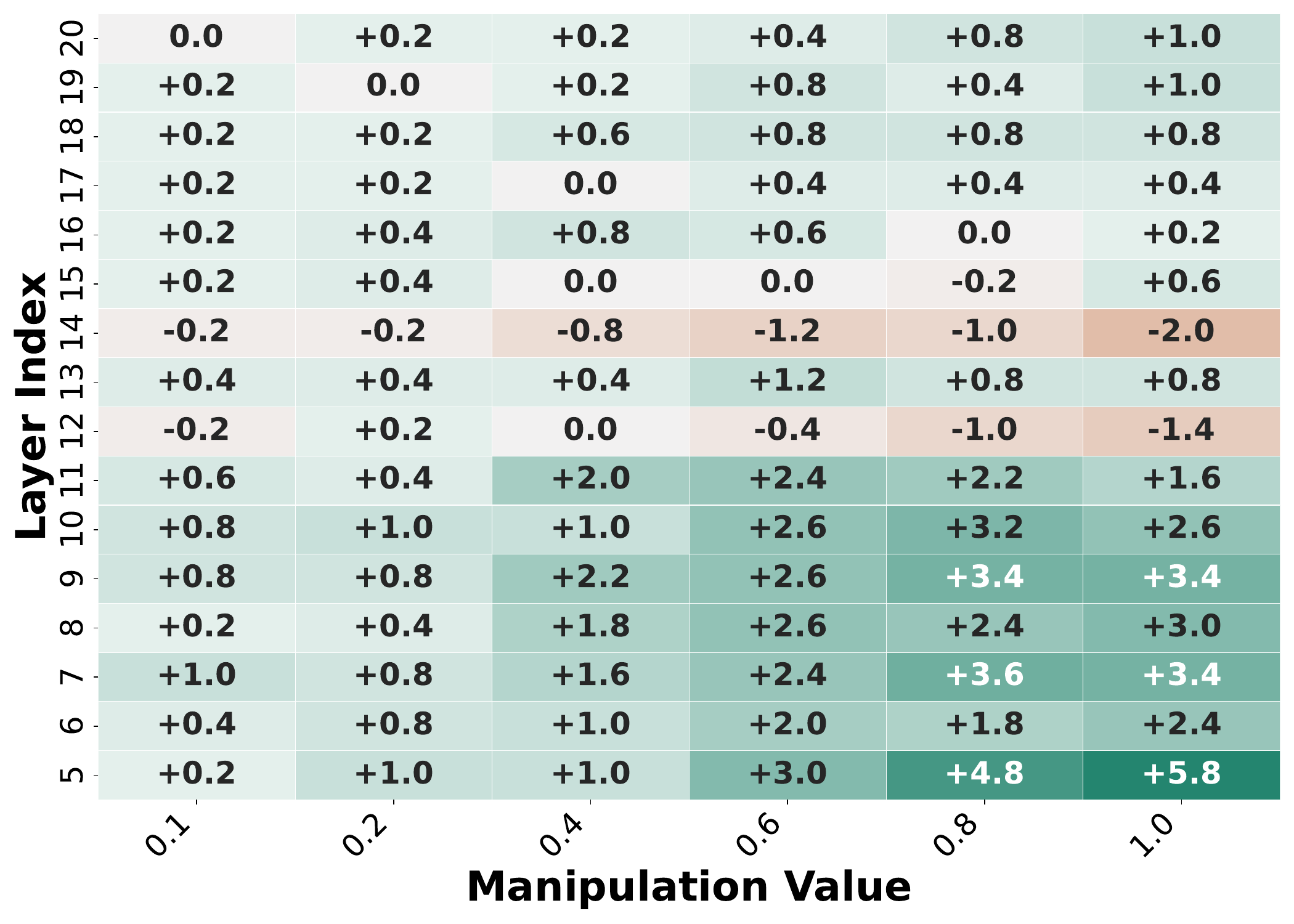}
        \caption{Grid search on \texttt{PaliGemma2-3B} to locate the best $(\ell^*, \alpha^*)$ for steering the model's spatial reasoning abilities. In this case, $\ell^* = 5$ and $\alpha^* = 1.0$.}
        % We show the grid search on the spatial relationship task with \texttt{PaliGemma2-3B} model.}
        \label{subfig:grid_search_pg3b}
    \end{subfigure}
    \caption{Efficient Grid Search with \texttt{PaliGemma2-3B} on the Spatial Relationship Task.}
    \label{fig:grid_search}
\end{figure}

\textbf{Grid Search. } Given a set of concept‐specific steering vectors \(v^{(\ell)}\) extracted as described in \S\ref{sec:methods}, we identify the optimal injection layer \(\ell\) and scale factor \(\alpha\) via a simple grid search on a held‐out ``grid search'' training split of CV‐Bench.  Concretely, for any model with $L$ layers, we grid search on layer indices $\mathcal I = \{\ell_1, \cdots, \ell_{|\mathcal{I}|}\}$ and steering strength $\mathcal{A} = \{\alpha_1, \dots, \alpha_{|\mathcal{A}|}\}$. For each \((\ell,\alpha)\in\mathcal{I}\times\mathcal{A}\), we intervene on the activations of the target tokens (image, text or both) at layer \(\ell\) as 
\[ h'_{\mathrm{target}}(\ell)
= h_{\mathrm{target}}(\ell) \;+\;\alpha\,v^{(\ell)}. \] 
We then evaluate the model’s task accuracy \(\mathrm{Acc}(\ell,\alpha)\) on the grid‐search split.  The optimal hyperparameters are chosen as 
$ (\ell^*,\alpha^*)
=\argmax_{\ell\in\mathcal{L},\;\alpha\in\mathcal{A}}
\mathrm{Acc}(\ell,\alpha)$. 
Once \((\ell^*,\alpha^*)\) is found, we fix these values for all subsequent evaluations on the held‐out test set.
In our experiments, we set $\mathcal{A}$ to $\{0.1,0.2,0.4,0.6,0.8,1.0\}$ if the steering vectors are unnormalized (e.g., those found via MeanShift). If vectors are normalized (for SAE and Probe), for \texttt{PaliGemma2-3B} and \texttt{Paligemma2-10B}, we set $\mathcal{A}$ to $\{10,20,30,40,50,60\}$, and for \texttt{Idefics3-8B-Llama3}, we use $\{0.2,0.4,0.6,0.8,1.0,1.2\}$. because the average norm of its hidden states is much smaller than that of the \texttt{PaliGemma2} models. We set $\mathcal{I}$ to be the middle layers, where we observe the learning from image tokens is predominantly happening (see Figure~\ref{subfig:efficient_grid_search}): $\{5, 6, \ldots, 20\}$ for \texttt{PaliGemma2-3B} and \texttt{Idefics3-8B-Llama3}, and $\{15, 16, \ldots, 30\}$ for \texttt{PaliGemma2-10B}.
Notably, we never steer the output token, as our focus is on modifying the model's internal representations.

\begin{figure}[t]
    \centering
    \begin{subfigure}{0.3\textwidth}
    \includegraphics[width=\linewidth]{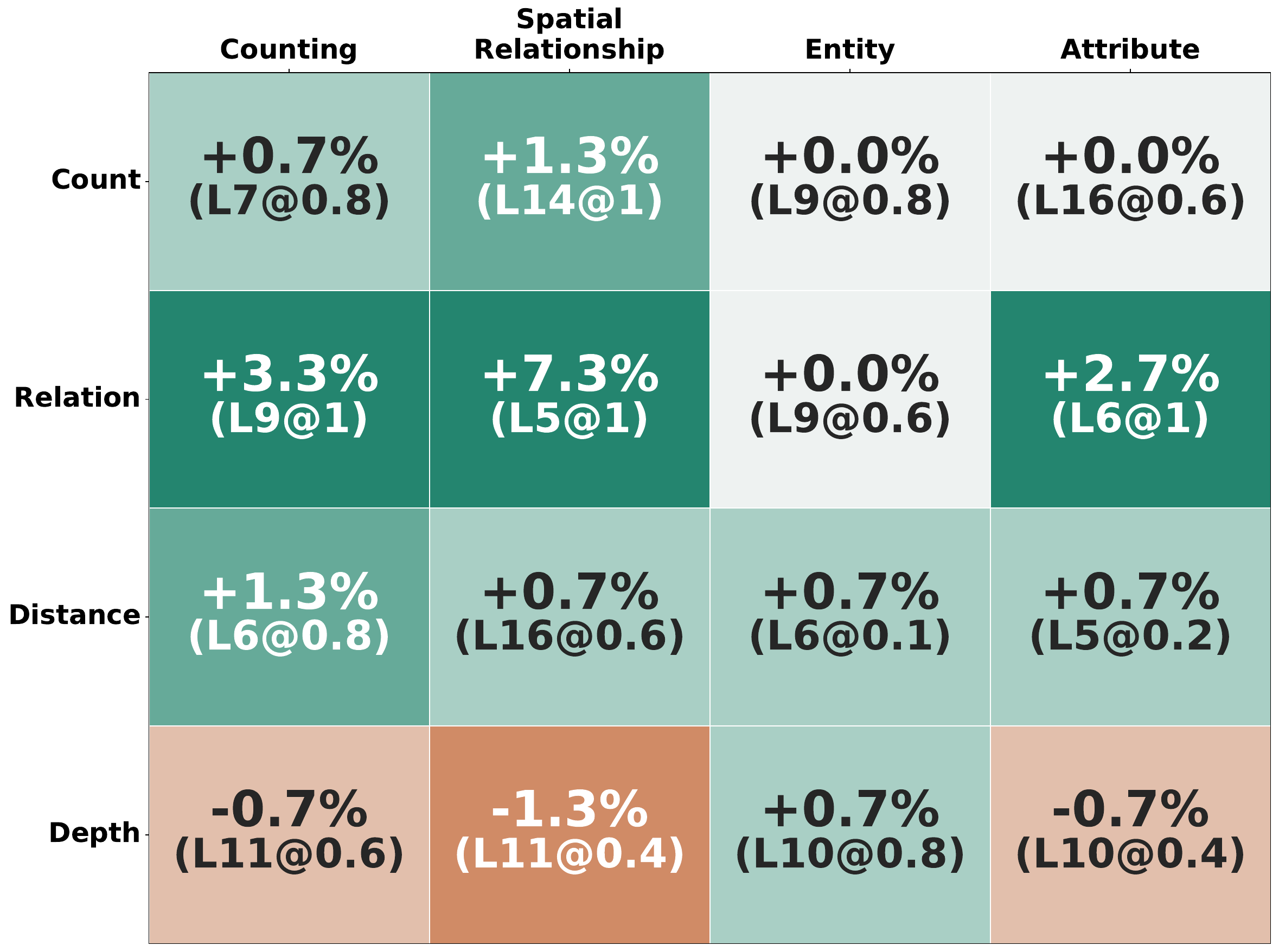}
    \caption{Intervening Text Tokens}
    \end{subfigure}
    \begin{subfigure}{0.3\textwidth}
    \includegraphics[width=\linewidth]{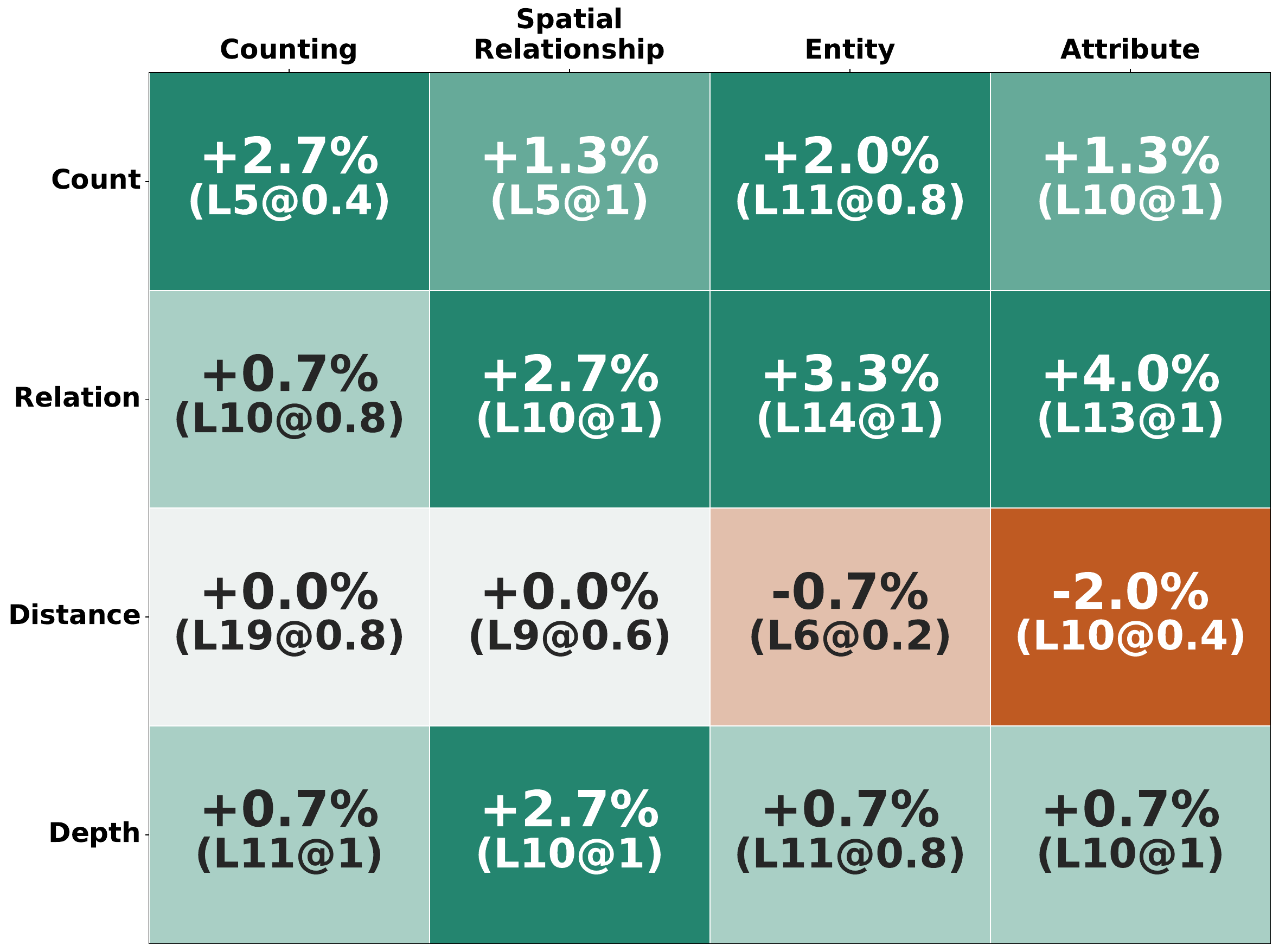}
    \caption{Intervening Image Tokens}
    \end{subfigure}
    \begin{subfigure}{0.3\textwidth}
    \includegraphics[width=\linewidth]{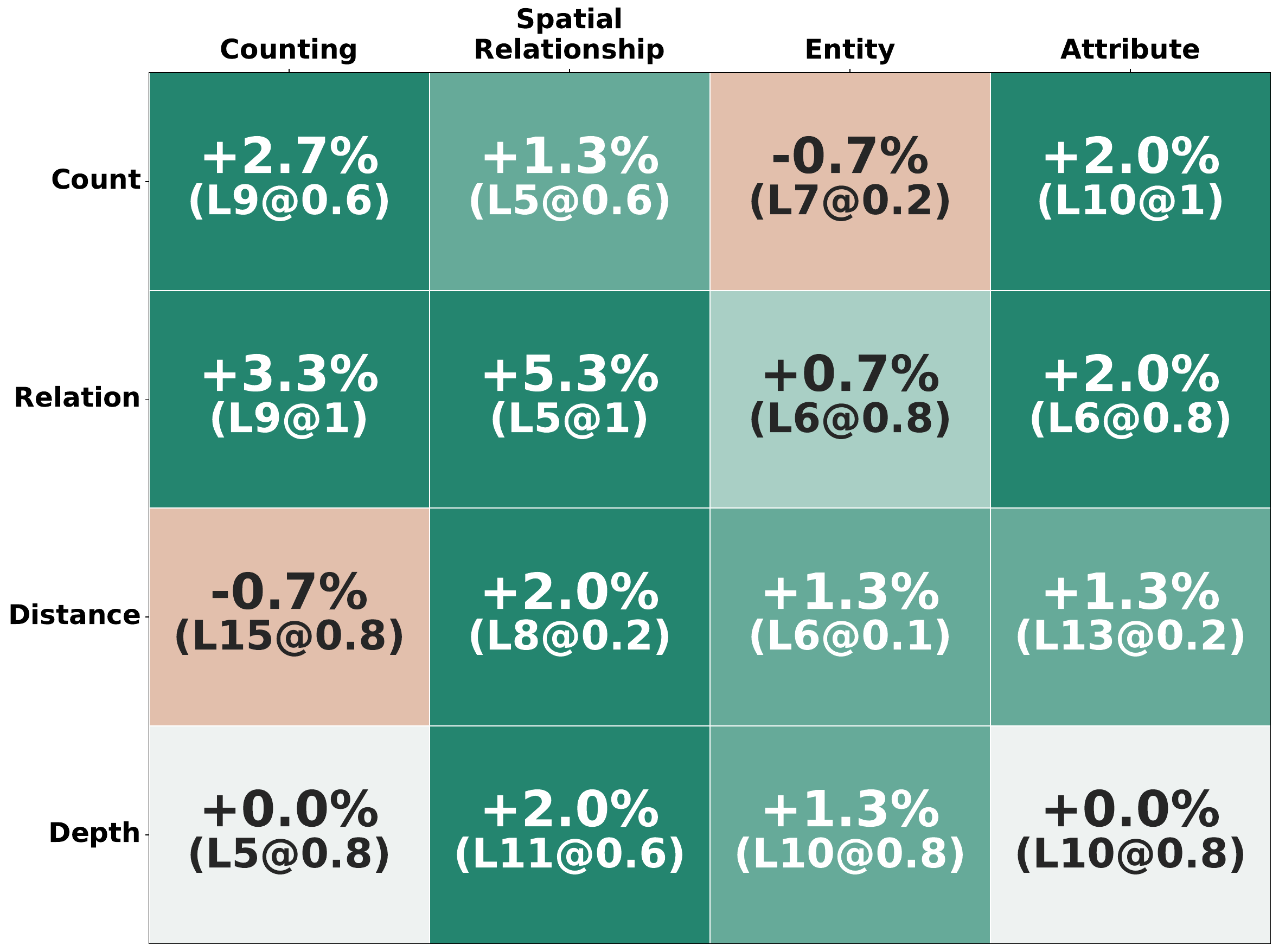}
    \caption{Intervening Both Tokens}
    \end{subfigure}
    \caption{Performance improvements on CV-Bench tasks when steering PaliGemma2-3B with MeanShift vectors. Each cell shows the percentage improvement in accuracy relative to the baseline. Rows represent different CV-Bench tasks, while columns represent different feature vectors used for steering. Text below improvements indicates the optimal layer number and intervention strength. }
    \label{fig:intervention_results_paligemma}
\end{figure}

\subsection{Results} \label{ssec:cvbench_results}

Table \ref{tab:combined_results} presents a comparative analysis of three different models, \texttt{PaliGemma2-3B}, \texttt{PaliGemma2-10B}, and \texttt{Idefics3-8B-Llama3}, on tasks related to spatial relationships and counting in CV-Bench. The performance is evaluated with and without intervention tokens (text, image, or both) and across different steering methods (SAE, Probe, MeanShift, and Prompting). 

\textbf{Prompting Barely Steers. } Table \ref{tab:combined_results} indicates that prompting is often less effective than targeted interventions using text and/or image tokens, and in some cases even deleterious for model performance (especially for spatial relation tasks). Interestingly, this finding deviates from the text-only observations of \textsc{AxBench} \citep{wu2025axbench}, most likely due to the multimodality of the models and tasks -- \emph{i.e.}, textual prompts are not sufficient for multimodal LLMs to better \textit{understand} images. 

\textbf{Steering Interventions Prove Effective.} Table \ref{tab:combined_results}   demonstrates that steering interventions, using text, image, or combined tokens, consistently improve model performance on spatial relationship and counting tasks over baseline levels. For instance, PaliGemma2-3B's ``Relation'' accuracy with MeanShift rose from 76.0 to 83.3 using both tokens, illustrating the general efficacy of these mechanisms.

\textbf{MeanShift Demonstrates Strong Overall Performance.} As shown in Table \ref{tab:combined_results}, Among the evaluated methods (SAE, Probe, MeanShift), MeanShift frequently performs as the most effective. This highlights MeanShift's robustness in leveraging textual representations and transferring to steer multimodal models, particularly for relational tasks.

\textbf{Steering More Impactful for Spatial Relationships.} Interventions yield more substantial accuracy improvements in the ``Spatial Relationship'' task than in ``Counting''. For instance, as shown in Table \ref{tab:combined_results}, with both tokens and MeanShift, PaliGemma2-3B gained +7.3 for relationships but only +2.7 for counting. This disparity may stem from spatial relationships being more directly influenced by highlighting salient object features and positions, while counting might demand a more holistic scene interpretation, less directly aided by these specific steering methods. 

\textbf{Intervention Transfers Across Tasks. } As shown in Figure \ref{fig:intervention_results_paligemma}, intervention using a feature $\mathcal{T}$ can sometimes be effective even when transferred to a different task $\mathcal{T}'$. For instance, intervention using the ``Attribute'' vector yields improvement in the ``Relation'' task, as does the ``Entity'' vector to a lesser extent. It may be that improved attribute and entity abilities aid the model in identifying the objects whose spatial relationship is being queried.
\section{Steering Improvements Generalize Out-of-Distribution} \label{sec:ood}

\begin{table}[t]
\centering
\NiceMatrixOptions{cell-space-top-limit=1pt,cell-space-bottom-limit=1pt}
\begin{adjustbox}{width=\textwidth}
\begin{NiceTabular}{l|l|l|lllll}[code-before = {\rowcolors{3}{gray!12}{white}}]
\toprule
\multirow{2}{*}{\centering\textsc{\textbf{Dataset}}} & 
\multirow{2}{*}{\centering\textsc{\textbf{\makecell[l]{Visual\\Concept}}}} & 
\multirow{2}{*}{\centering\textsc{\textbf{Model}}} & 
\multicolumn{5}{c}{\textsc{\textbf{Intervention Method}}} \\
& & & \textsc{Baseline} & \textsc{Prompting} & \textsc{SAE} & \textsc{Probe} & \textsc{Meanshift} \\
\midrule
% ---------------- Whatsup-A ----------------
\multirow{4}{*}{\texttt{What'sUp-A}} & 
\multirow{4}{*}{\texttt{\makecell[l]{Spatial\\Relation}}} & 
\texttt{PaliGemma2-3B} & 
\score{62.7}{4.2} & 
\score{59.4}{}{-3.3} &
\score{71.8}{3.7}{+9.1} & 
\score{78.5}{3.4}{+15.8} & 
\score{75.4}{3.5}{+12.7} \\
& & \texttt{PaliGemma2-10B} & 
\score{68.5}{3.9} & 
\score{71.0}{}{+2.5} &
\score{80.1}{3.3}{+11.6} & 
\score{71.6}{3.7}{+3.0} & 
\score{74.9}{3.6}{+6.4} \\
& & \texttt{Idefics3-8B-Llama3} & 
\score{62.2}{4.2} & 
\score{63.5}{}{+1.3} &
\score{64.1}{4.1}{+1.9} & 
\score{62.2}{4.2}{+0.0} & 
\score{61.9}{4.2}{-0.3} \\
& & \textsc{Average Improvement} & 
{--} & 
{+0.2} &
{\underline{\textbf{+7.6}}} & 
{+6.3} & 
{+6.3} \\
\midrule
% ---------------- Whatsup-B ----------------
\multirow{4}{*}{\texttt{What'sUp-B}} & 
\multirow{4}{*}{\texttt{\makecell[l]{Spatial\\Relation}}} & 
\texttt{PaliGemma2-3B} & 
\score{60.6}{} & 
\score{58.4}{}{-2.2} &
\score{58.9}{}{-1.7} & 
\score{57.5}{}{-3.1} & 
\score{60.3}{}{-0.3} \\
& & \texttt{PaliGemma2-10B} & 
\score{81.8}{} & 
\score{81.5}{}{-0.3} &
\score{82.4}{}{+0.6} & 
\score{82.1}{}{+0.3} & 
\score{82.1}{}{+0.3} \\
& & \texttt{Idefics3-8B-Llama3} & 
\score{52.0}{} & 
\score{56.9}{}{+4.9} &
\score{56.2}{}{+4.2} & 
\score{57.0}{}{+5.0} & 
\score{63.4}{}{+11.5} \\
& & \textsc{Average Improvement} & 
{--} & 
{+0.8} &
{+1.0} & 
{+0.8} & 
{\underline{\textbf{+3.8}}} \\
\midrule
% ---------------- Blink-object_localization ----------------
\multirow{4}{*}{\texttt{\makecell[l]{BLINK\\Object Localization}}} & 
\multirow{4}{*}{\texttt{\makecell[l]{Spatial\\Relation}}} & 
\texttt{PaliGemma2-3B} & 
\score{41.2}{} & 
\score{42.3}{}{+1.0} &
\score{43.3}{}{+2.1} & 
\score{42.3}{}{+1.0} & 
\score{44.3}{}{+3.1} \\
& & \texttt{PaliGemma2-10B} & 
\score{51.6}{} & 
\score{51.6}{}{+0.0} &
\score{54.6}{}{+3.1} & 
\score{53.6}{}{+2.1} & 
\score{57.7}{}{+6.2} \\
& & \texttt{Idefics3-8B-Llama3} & 
\score{53.6}{} & 
\score{54.6}{}{+1.0} &
\score{56.7}{}{+3.1} & 
\score{53.6}{}{+0.0} & 
\score{55.7}{}{+2.1} \\
& & \textsc{Average Improvement} & 
{--} & 
{+0.7} &
{+2.8} & 
{+1.0} & 
{\underline{\textbf{+3.8}}} \\
\midrule
% ---------------- CLEVR normal ----------------
\multirow{4}{*}{\texttt{CLEVR}} & 
\multirow{4}{*}{\texttt{Count}} & 
\texttt{PaliGemma2-3B} & 
\score{52.4}{} & 
\score{55.1}{}{+2.7} &
\score{70.7}{}{+18.2} & 
\score{56.4}{}{+4.0} & 
\score{67.1}{}{+14.7} \\
& & \texttt{PaliGemma2-10B} & 
\score{70.7}{} & 
\score{70.7}{}{+0.0} &
\score{74.9}{}{+4.2} & 
\score{71.6}{}{+0.9} & 
\score{80.4}{}{+9.8} \\
& & \texttt{Idefics3-8B-Llama3} & 
\score{59.8}{} & 
\score{65.1}{}{+5.3} &
\score{88.0}{}{+28.2} & 
\score{84.4}{}{+24.7} & 
\score{94.0}{}{+34.2} \\
& & \textsc{Average Improvement} & 
{--} & 
{+2.7} &
{+16.9} & 
{+9.9} & 
{\underline{\textbf{+19.6}}} \\
\midrule
% ---------------- CLEVR challenging ----------------
\multirow{4}{*}{\texttt{Super-CLEVR}} & 
\multirow{4}{*}{\texttt{Count}} & 
\texttt{PaliGemma2-3B} & 
\score{26.9}{} & 
\score{28.0}{}{+1.1} &
\score{32.0}{}{+5.1} & 
\score{30.3}{}{+3.4} & 
\score{33.1}{}{+6.3} \\
& & \texttt{PaliGemma2-10B} & 
\score{40.0}{} & 
\score{44.0}{}{+4.0} &
\score{40.6}{}{+0.6} & 
\score{40.0}{}{+0.0} & 
\score{44.6}{}{+4.6} \\
& & \texttt{Idefics3-8B-Llama3} & 
\score{66.5}{} & 
\score{64.0}{}{-2.5} &
\score{66.5}{}{+0.0} & 
\score{67.5}{}{+1.0} & 
\score{68.5}{}{+2.0} \\
& & \textsc{Average Improvement} & 
{--} & 
{+0.9} &
{+1.9} & 
{+1.5} & 
{\underline{\textbf{+4.3}}} \\
\midrule
% Averaging
\multicolumn{3}{c}{\textsc{Average Improvement}} & 
{--} & 
{+1.0} &
{+6.0} &
{+3.9} &
{\underline{\textbf{+7.6}}} \\
\bottomrule
\end{NiceTabular}
\end{adjustbox}
\vspace{1mm}
\caption{Performance of textual steering on out-of-distribution datasets.}
\label{tab:dataset_performance_comparison}
\end{table}

We now examine the ability of textual steering methods for MLLMs to generalize out-of-distribution, \emph{i.e.}, to datasets on which the steering method's hyperparameters $(\ell, \alpha)$ have not been tuned. 

\subsection{Setup}
\textbf{Datasets. } We consider the transferability of textual steering on five datasets: What'sUp-A, What'sUp-B, BLINK Object Localization, CLEVR, and Super-CLEVR. What'sUp-A contains 408 images of pairs of household objects arranged in clear spatial relations of \{``on'', ``under'', ``left'', and ``right''\}, while What'sUp-B similarly contains 412 images with objects in the image closer in size \citep{kamath2023s}. The BLINK Object Localization category contains 122 questions related to bounding boxes for large objects \citep{fu2024blink}. Finally, we sampled 500 datapoints from CLEVR \citep{johnson2017clevr} and 200 datapoints from Super-CLEVR \citep{Li_2023_CVPR} to evaluate the OOD accuracy of textual steering in counting.

\textbf{Steering Vector Hyperparameter Selection.} We examine the previous three steering methodologies---SAE, MeanShift, and Probe---with a single choice of layer $\ell$ and scale factor $\alpha$ chosen independently of the test dataset. Specifically, for each test dataset, we select the $(\ell, \alpha)$ pair that performed best on the corresponding CV-Bench task category (\textit{e.g.}, ``Relation'' for the What'sUp datasets and Blink Object Localization focusing on spatial relationships, and ``Count'' for CLEVR and Super-CLEVR). 

We emphasize that the steering methods' hyperparameters are \emph{not} tuned to the datasets considered in this section, making this a true test of out-of-distribution generalization. Similarly, our prompting baseline uses the exact prompt prefix that performed best on the associated CV-Bench tasks. The only adaptation made was the use of a small validation subset (50 datapoints for What'sUp and CLEVR, 25 datapoints for BLINK Object Localization and Super-CLEVR) to determine the most effective token type for intervention (image, text, or both) before evaluating on the remaining data.

\subsection{Results}
\textbf{Steering Remains Broadly Effective. } Table~\ref{tab:dataset_performance_comparison} demonstrates that textual interventions are broadly effective across all 5 datasets considered, attaining an average improvement over all models and datasets of at least +3.9\% for all three vector-based steering methods. Prompting, meanwhile, averaged a +1.0\% improvement and worsened model performance in 4 cases, suggesting that it may be less effective for MLLMs than for text-only LLMs \citep{wu2025axbench}. Notably, the out-of-distribution performance of textual steering appears to surpass its in-distribution performance on CV-Bench. We hypothesize that this may be due to the fact that the What'sUp and CLEVR datasets rely ``purely'' upon spatial reasoning and counting, respectively, whereas CV-Bench tests broader compositional understanding and thus benefits less from any individual intervention. 

% \jcomment{Discuss that these results are even better than in-distribution, right? Because datasets are easier than CV Bench?}\wgcomment{I feel like it is because What'sUp and CLVER are more pure on testing spatial relationships and counting, and cvbench contains more compositional understanding}
% Prompting, of course, remains ineffective, as there is no distinction between ``in-distribution'' and ``out-of-distribution'' prompting, \emph{i.e.}, our prompting methodology is not tuned to the underlying dataset. 

\textbf{MeanShift Often Most Effective.} We also find that MeanShift attains the greatest average performance of +7.6\% across all models and datasets, and attains the best performance on a majority of all model-dataset pairs. This extends our finding from Section~\ref{sec:steering_improves}, suggesting that MeanShift may be the most broadly effective textual steering method for MLLMs.

\textbf{Pronounced Gains For CLEVR. } The dataset on which the largest improvements in performance are observed is CLEVR, by a 3x factor. The improvements of \t{Idefics3-8B-Llama3} largely drive these gains, with all vector-based steering methods (\emph{i.e.,} non-prompting) achieving improvements of at least +24\%. In contrast, Super-CLEVR achieves an accuracy improvement of only +4.3\%, perhaps simply due to the fact that a more difficult dataset is more difficult to improve upon. Curiously, however, the baseline \t{Idefics3-8B-Llama3} model achieves a \emph{higher} accuracy on this dataset than on CLEVR. 

\section{Discussion} \label{sec:discussion}

We examine the ability of multimodal large language models (MLLMs) derived from backbone (text-only) large language models to be modified using textual steering vectors from their text-only backbone. We find that MeanShift, sparse autoencoders (SAEs), and linear probes can all enhance MLLMs' visual reasoning across diverse tasks on CV-Bench, with MeanShift demonstrating the strongest overall performance. Notably, the steering vectors derived from CV-Bench also generalize out-of-distribution to other datasets such as What'sUp, BLINK, and CLEVR, underscoring text-driven steering as a powerful and efficient medium for enhancing visual reasoning in MLLMs. 
%\jedit{
A primary limitation of our work is that it requires the MLLM to have a text-only backbone, and that its efficacy is contingent upon the quality of the textual steering vectors derived from the backbone (which may be imperfect, see e.g.\ \citep{wu2025axbench,heap2025sparse}).
%} 
An interesting direction for future study is to further investigate and contrast the possibility of steering vectors extracted ``directly'' from MLLMs, rather than from their text-only backbones. 
% \jcomment{Mention limitations for the sake of the checklist?} \dfcomment{TODO: add limitations in the discussion. }

\section*{Acknowledgments}
This research is supported in part by AWS credits through an Amazon Faculty research award, a NAIRR Pilot award, and Microsft accelerating foundations research grant.
R.\ Jia was also supported by the National Science Foundation under Grant No.\ IIS-2403436. 
V.\ Sharan was supported in part by an NSF CAREER Award CCF-2239265 and an Amazon Research Award.
W.\ Neiswanger was supported in part by the National Science Foundation under Grant No.\ CMMI-2427856.
J.\ Asilis was supported by the National Science Foundation Graduate Research Fellowship
Program under Grant No.\ DGE-1842487.
Any opinions, findings, and conclusions or recommendations expressed in this material are those of the author(s) and do not necessarily reflect the views of the National Science Foundation.
The work was done in part while some of the authors were visiting the Simons Institute for the Theory of Computing. D.\ Fu would like to thank Muru Zhang and Yuqing Yang for meaningful discussions on probing multimodal LLMs.

\clearpage
\bibliographystyle{plainnat}
\bibliography{main}

\clearpage
\appendix
\section*{Appendix}
\startcontents[appendix]
\addcontentsline{toc}{chapter}{Appendix}
\renewcommand{\thesection}{\Alph{section}} 
\printcontents[appendix]{}{1}{\setcounter{tocdepth}{3}}
\setcounter{section}{0}

\section{Steering Vector Methodology}
\subsection{Sparse Autoencoders}
\label{Appendix:SAEs}

We now provide further detail regarding the extraction of textual steering vectors for visual concepts using SAEs. 

Recall that we consider four important taxonomies for image-related concepts: spatial relationship, counting, attribute, and entity. For each taxonomy, we sample $K$ sentences $\{s_1, \ldots, s_K\}$ containing such visual concepts. In practice, we set $K$ to 20. For each sentence $s_j$, we identify the anchor word for such visual concept as $w_j$, thus forming sentence-anchor pairs $(s_j, w_j)$. See table~\ref{tab:taxonomy_sample} for several examples. 

\begin{table}[h]
    \centering
    \caption{Sample sentence and anchor word pairs for various taxonomies.}
    \small
    \begin{tabular}{l|l|l}
        \textsc{Taxonomy} & \textsc{Sentence $s_j$} & \textsc{Anchor Word $w_j$}  \\
        \midrule
        \multirow{2}{*}{Spatial Relationship} & The cat is on the table & on \\
            & She put the book under the chair & under \\
        \midrule
        \multirow{2}{*}{Counting} &There are three apples in the basket & three \\ 
            &The teacher counted five children & five \\ 
        \midrule
        \multirow{2}{*}{Attribute}  & The red car stopped at the light & red \\ 
             & She wore a beautiful dress & beautiful \\ 
        \midrule
        \multirow{2}{*}{Entity}  & The dog barked at the mailman & dog \\ 
             & A tree fell during the storm & tree \\ 
    \end{tabular}
    
    \label{tab:taxonomy_sample}
\end{table}

\begin{figure}[h]
\centering
\tcbset{colback=bboxbg}
\begin{tcolorbox}[
    enhanced,
    %breakable,
    skin=standard,
    % colback=bboxbg,
    % colframe=black,
    boxrule=0.5pt,
    % sharp corners,
    before upper=\setlength{\parskip}{0pt}\setlength{\parindent}{0pt},
    segmentation style={draw=none,fill=none},
    title=\textsc{Feature Alignment Verification},
    % label={box:steering_prompt},
    % caption={Example system and user prompt template for generating steering instructions}
    ]
\t{Task: Determine if a neural network's sparse autoencoder (SAE) feature aligns with the taxonomy "\{taxonomy\}". \\ \\
Taxonomy Definition: \{taxonomy\_definition\} \\ \\
Feature Information: \\
1. Feature's explanation: \{feature\_explanation\} \\
2. Top activation examples (tokens wrapped in <top>...</top> have the highest activation values and are the most important to focus on):\\
\hspace*{1em} 1. \{activation\_example\_1\} \\
\hspace*{1em} 2. \{activation\_example\_2\} \\
\hspace*{1em} 3. \{activation\_example\_3\} \\
\hspace*{1em} 4. \{activation\_example\_4\} \\
\hspace*{1em} 5. \{activation\_example\_5\} \\ \\
Examples of features that DO align with the \{taxonomy\} taxonomy (notice how the key words are highlighted with <top>...</top> tags): \\
Example 1: \\
- Explanation: \{explanation\_1\} \\ 
- Activations: \{activations\_1\} \\ 
Example 2: \\
- Explanation: \{explanation\_2\} \\ 
- Activations: \{activations\_2\} \\ \\
When making your decision, you should follow these rules: \\
1. First pay attention to the feature's explanation. \\
2. If you cannot decide, you should then pay special attention to the tokens highlighted with <top>...</top> tags, as these are the most highly activated tokens and strongest indicators of what the feature detects. \\
3. Also consider the diversity of the activation examples provided. If one feature only activates one particular word, it may not be as aligned as a feature that activates on a variety of words. \\ \\
Based on the feature's explanation and the highlighted tokens in the activation examples, does this feature specifically detect or respond to \{taxonomy\_definition\}? Your answer should start with YES or NO, then provide a brief reason. Do not start with any other words or phrases such as `answer'.
}
\end{tcolorbox}
% \tcbset{reset}
\caption{Prompt template for querying \t{GPT-o3-mini} to verify whether a given feature is related to a visual taxonomy. For each taxonomy, the template employs a brief definition of the taxonomy, two example features that align with each taxonomy (for few-shot learning), and the top five activations of the feature in question.}
\label{box:o3_prompt}
\end{figure}

\subsection{Prompting}
\label{Appendix:prompting}

We now elaborate upon our generation of prompts for eliciting taxonomy-specific visual reasoning in MLLMs. As described in Section~\ref{ssec:prompting}, we generate a total of 30 candidate prompts for each taxonomy $\mathcal{T}$. To do so, we use template shown in figure~\ref{box:steering_prompt}. Here, we set the \texttt{num instructions} to $3$ and \texttt{word count} $\in$ \(\{5,10,15,20,25,30,35,40,45,50\}\), resulting in total $3\times10=30$ steering prompts.

\begin{figure}[h]
\centering
\tcbset{colback=bboxbg}
\begin{tcolorbox}[
    enhanced,
    %breakable,
    % float,
    title=\textsc{Steering Prompt Generation},
    % label={box:steering_prompt},
    % caption={Example system and user prompt template for generating steering instructions}
    ]
\textbf{System prompt:}
\t{You are an expert at creating concise, clear instructions for Multimodal Large Language Models (MLLM). \\ \\ 
Your task: \\
- Generate \{num\_instructions\} different instruction(5) that will make the Model focus on \{concept\} when answering questions about images \\
- Each instruction must be within \{word\_count\} words \\
- Instructions should be direct and actionable, focusing specifically on how to emphasize \{concept\} \\ \\
IMPORTANT FORMAT REQUIREMENTS: \\
- Begin each instruction with "INSTRUCTION:" followed by the instruction text \\
- Put each instruction on its own line \\
- Do not include any numbering, bullets, or other text beyond the requested instructions \\
- Do not include any explanations, introductions, or conclusions \\ \\
Example format for 2 instructions: \\
INSTRUCTION: First instruction text here within word limit. \\
INSTRUCTION: Second instruction text here within word limit. \\ \\
}

\textbf{User prompt:}
\t{Create \{num\_instructions\} instruction(s) about \{concept\} using \{word\_count\} words or fewer each.}
\end{tcolorbox}
% \tcbset{reset}
\caption{System and user prompt template for generating MLLM prompts.}
\label{box:steering_prompt}
\end{figure}

% {
% \tcbset{
%     enhanced, 
%     frame hidden,
%     breakable,
%     attach boxed title to top left={
%     xshift=0.5cm,
%     yshift=-\tcboxedtitleheight/2},
%     top=4mm,
%     coltitle=bboxbg,
%     beforeafter skip=\baselineskip,
% }
% \tcbset{colback=bboxbg}
% \begin{tcolorbox}[title=\textsc{TITLE}]
% %\includegraphics[width=0.5\linewidth]{figures/ex_93_img_0.png}

% \textbf{System prompt:}
% \t{You are an expert at creating concise, clear instructions for Multimodal Large Language Models (MLLM). \\ \\ 
% Your task: \\
% - Generate \{num\_instructions\} different instruction(5) that will make the Model focus on \{concept\} when answering questions about images \\
% - Each instruction must be within \{word\_count\} words \\
% - Instructions should be direct and actionable, focusing specifically on how to emphasize \{concept\} \\ \\
% IMPORTANT FORMAT REQUIREMENTS: \\
% - Begin each instruction with "INSTRUCTION:" followed by the instruction text \\
% - Put each instruction on its own line \\
% - Do not include any numbering, bullets, or other text beyond the requested instructions \\
% - Do not include any explanations, introductions, or conclusions \\ \\
% Example format for 2 instructions: \\
% INSTRUCTION: First instruction text here within word limit. \\
% INSTRUCTION: Second instruction text here within word limit. \\ \\
% }

% \textbf{User prompt:}
% \t{Create \{num\_instructions\} instruction(s) about \{concept\} using \{word\_count\} words or fewer each.}
% \end{tcolorbox}
% \tcbset{reset}
% }
\clearpage
\section{Additional Color Perception Intervention Examples}
\label{appendix:color_examples}

To further demonstrate the effectiveness of textual steering vectors in modifying visual understanding within MLLMs, we present additional color perception intervention examples using the same methodology described in \S\ref{sec:toy}.

\begin{figure}[h]
    \centering
    \begin{subfigure}[t]{0.48\textwidth}
        \includegraphics[width=\linewidth]{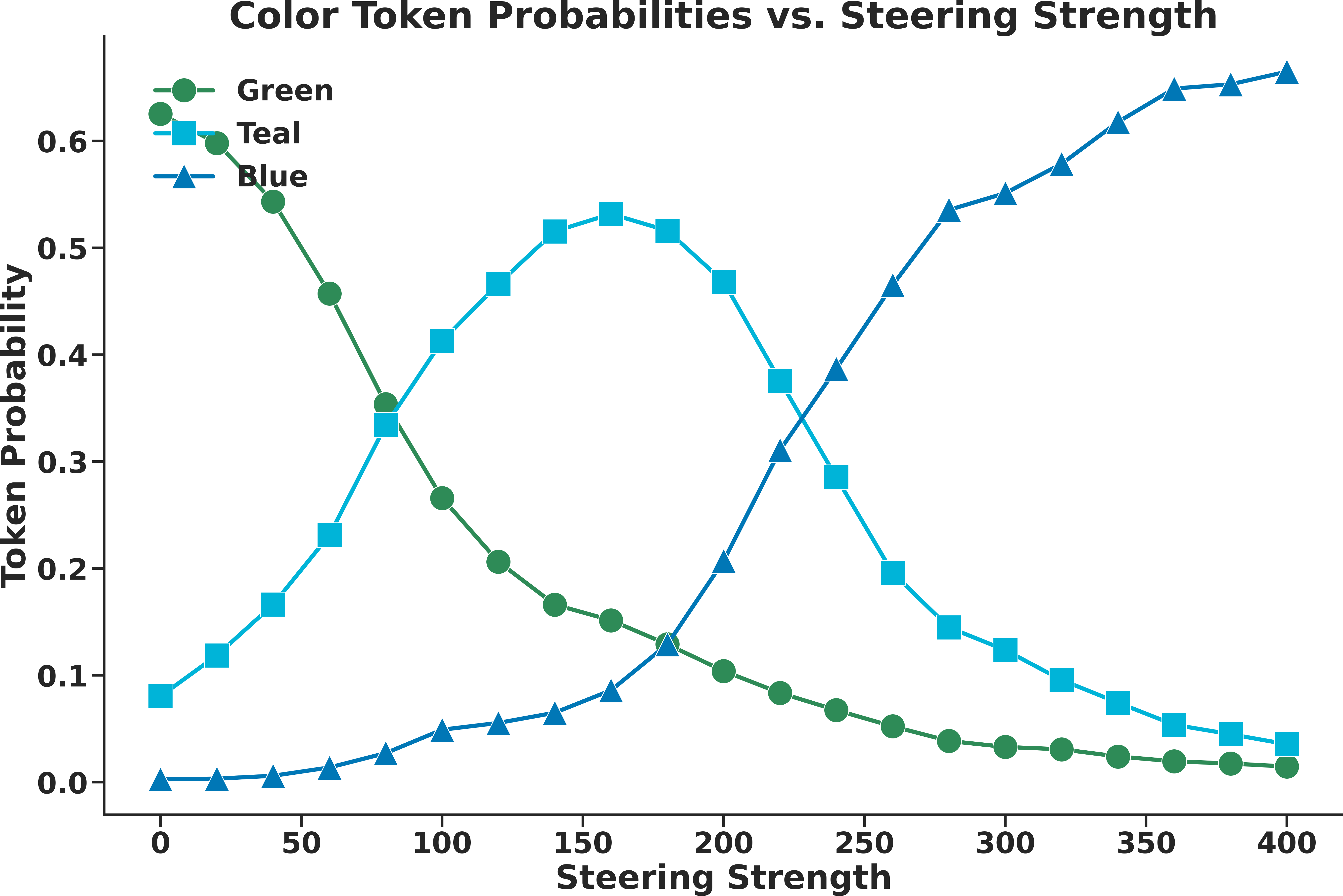}
        \caption{Steering a green image toward blue perception. As the scale factor increases, the model's interpretation shifts from green to teal, and ultimately to blue.}
        \label{fig:green_to_blue}
    \end{subfigure}
    \hfill
    \begin{subfigure}[t]{0.48\textwidth}
        \includegraphics[width=\linewidth]{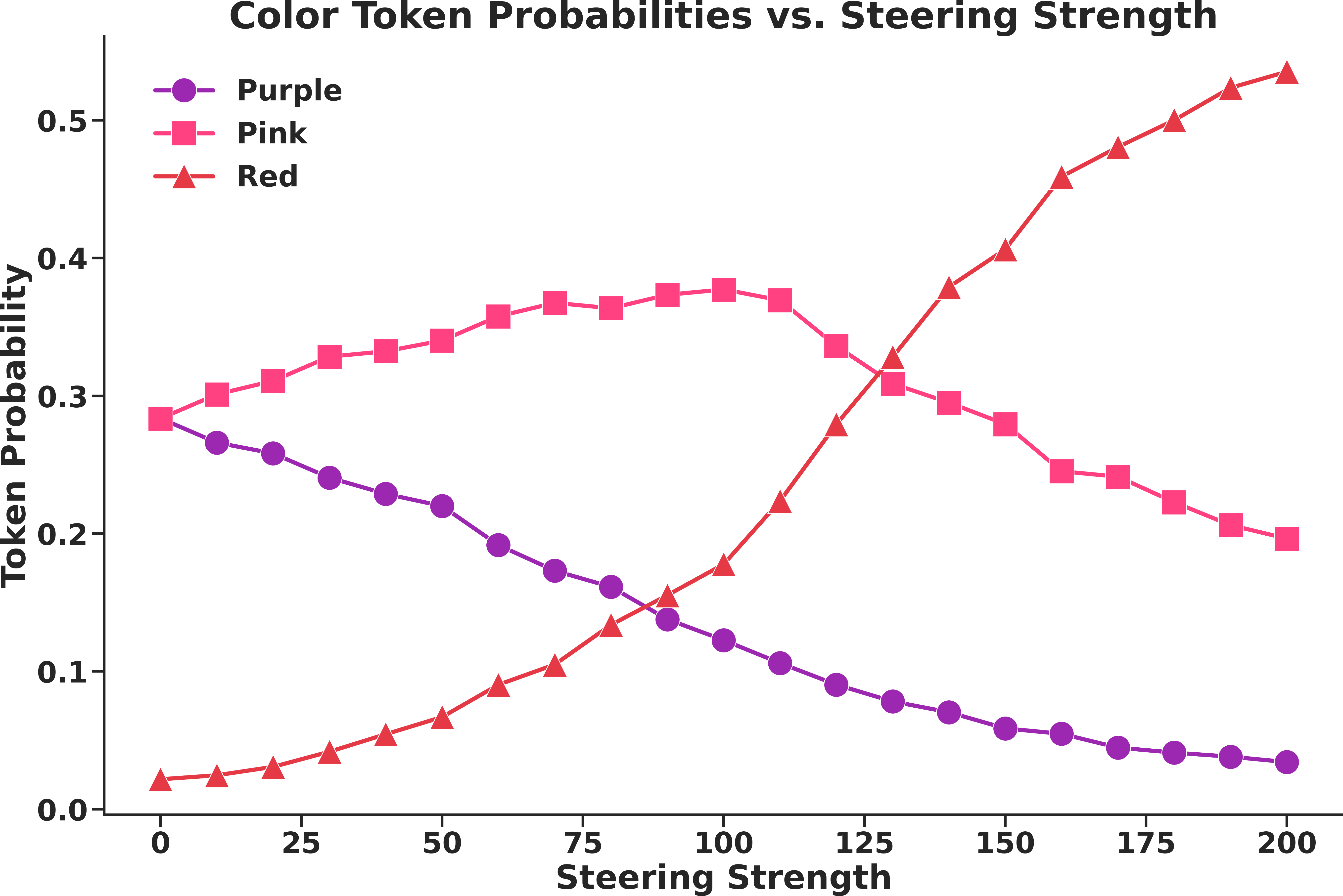}
        \caption{Steering a purple image toward red perception. The intervention gradually shifts the model's color association from purple to pink, and finally to red.}
        \label{fig:purple_to_red}
    \end{subfigure}
    
    \vspace{0.5cm}
    
    \begin{subfigure}[t]{0.48\textwidth}
        \includegraphics[width=\linewidth]{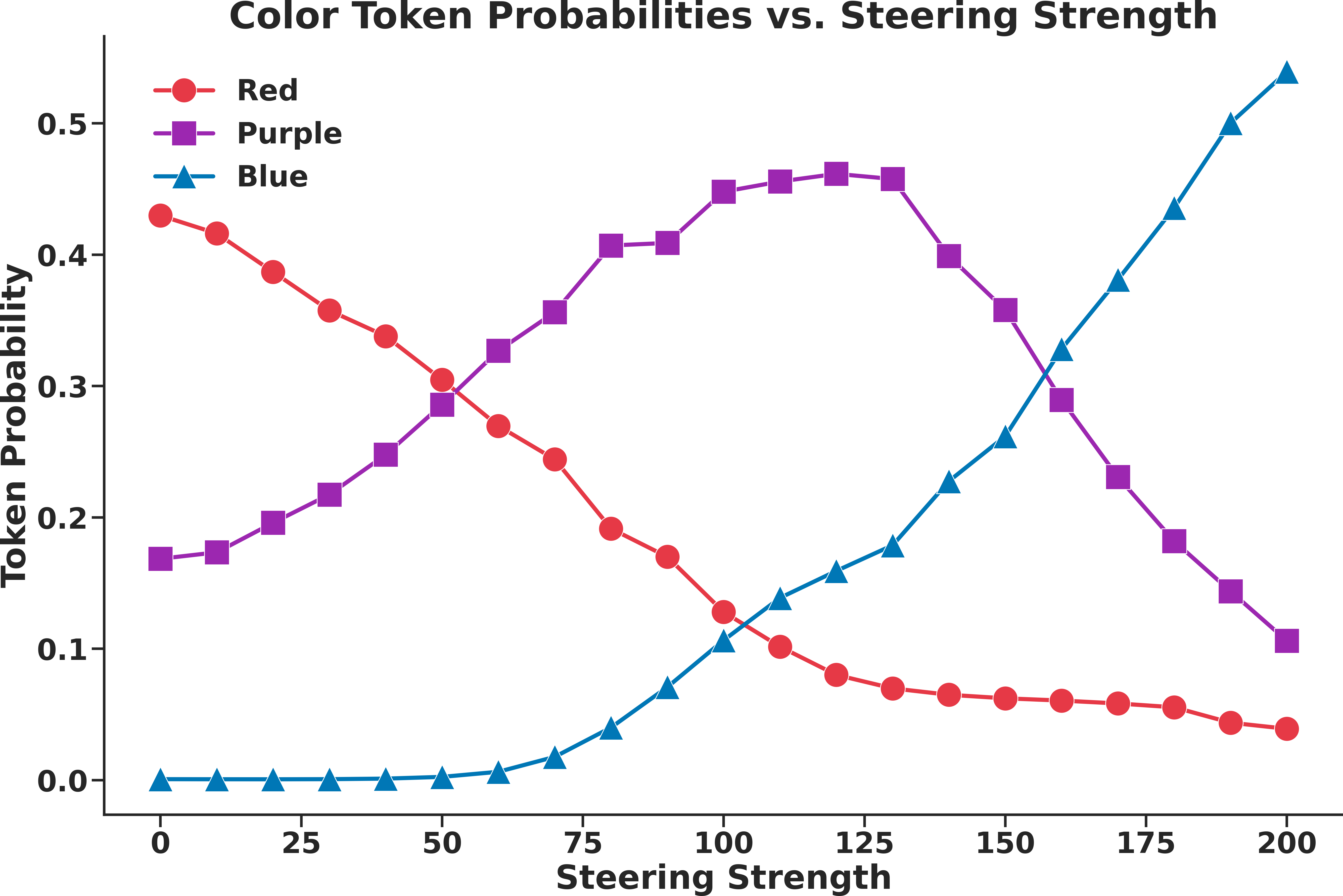}
        \caption{Steering an red image toward blue perception. The intervention causes a gradual shift from red to purple, and ultimately to blue.}
        \label{fig:red_to_blue}
    \end{subfigure}
    \hfill
    \begin{subfigure}[t]{0.48\textwidth}
        \includegraphics[width=\linewidth]{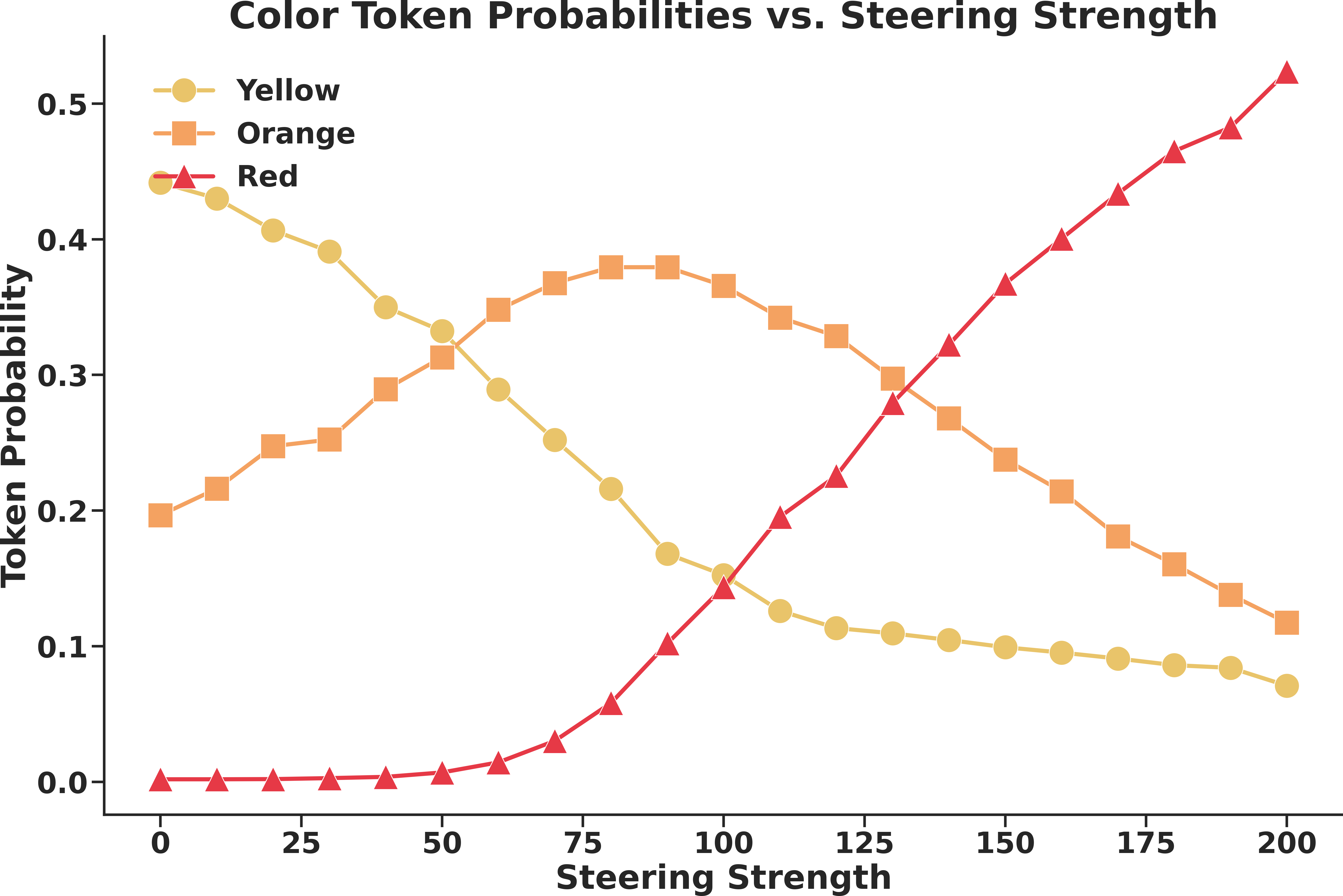}
        \caption{Another example of steering a yellow image toward red perception, using a different steering vector from layer 18 of \texttt{PaliGemma2-10B}. As the scale factor increases, the model's interpretation transitions from yellow to orange, and finally to red.}
        \label{fig:yellow_to_red2}
    \end{subfigure}
    
    \caption{Additional color perception intervention examples. In each case, we apply the normalized textual steering vector for the target color to the image tokens with increasing scale factors. The steering vectors are extracted from and applies to one selected layer from layer 17 to 20 in \texttt{PaliGemma2-10B}. The plots show token probability shifts, demonstrating how textual steering vectors can systematically modify the model's visual perception.}
    \label{fig:additional_color_examples}
\end{figure}

These additional examples further support our findings in \S\ref{sec:toy}. In each case, we see a clear progression of perception as the steering strength increases, with intermediate colors appearing during the transition. This confirms that textual steering vectors can produce predictable and continuous modifications to visual understanding.

Notably, all these interventions were performed using steering vectors derived solely from text data, yet they effectively modulate multimodal understanding. This provides additional evidence for our hypothesis that MLLMs develop unified cross-modal representations that can be manipulated through textual steering.
\clearpage
\section{Dataset Evaluation Details} \label{app:expt_details}
In this section, we explain in detail how we prompt and evaluate the model's performance across datasets and provide representative examples for each dataset.
Each prompt consists of four components: \texttt{model prefix}, \texttt{task prefix}, \texttt{taxonomy prefix}, and \texttt{question}. The \texttt{model prefix} is the specific instruction token sequence required by different model families to perform certain tasks. For \texttt{PaliGemma2} models, we use "\texttt{answer en}" as the model prefix, indicating that the model should answer in English for visual question answering tasks. For \texttt{Idefics3-8B-Llama3}, no model prefix is required, so this component remains empty.
The \texttt{task prefix} provides task-specific instructions that constrain the format of the model's response. In multiple-choice questions, we use a task prefix such as "\texttt{Answer the multiple choice question by only responding with the letter of the correct answer.}" In \texttt{CLEVR} abd \texttt{Super-CLEVR} counting questions, we use "\texttt{Answer the question by only responding the number.}"
The \texttt{taxonomy prefix} of each taxonomy is the prompt we sampled in Section~\ref{ssec:prompting}, and it is only non-empty for the \texttt{Prompt} method. The \texttt{question} component contains the original question format from the dataset.
Below are examples illustrating our prompting approach for each dataset.
\begin{figure}[ht]
\centering
\tcbset{colback=bboxbg}
\begin{tcolorbox}[title=\textsc{CV-Bench Relation}]
\ttfamily
Image: 
\includegraphics[width=0.5\linewidth]{figures/dataset_examples/cvbench1375.jpg}
\newline
{\color{blue}\textbf{[Model Prefix]} answer en} {\color{Pos}\textbf{[Task Prefix]} Answer the multiple choice question by only responding the letter of the correct answer.} {\color{red}\textbf{[Taxonomy Prefix]} Emphasize objects' positions relative to each other.} {\color{black}\textbf{[Question]} Considering the relative positions of the fork and the cup in the image provided, where is the fork located with respect to the cup? Select from the following choices.\newline(A) left\newline(B) right}
\end{tcolorbox}
\tcbset{reset}
\caption{Example prompt for the CV-Bench Relation dataset.}
\label{fig:cvbench_relation_example}
\end{figure}
\begin{figure}[ht]
\tcbset{colback=bboxbg}
\begin{tcolorbox}[title=\textsc{CV-Bench Count}]
\ttfamily
Image: 
\includegraphics[width=0.5\linewidth]{figures/dataset_examples/cvbench315.jpg}
\newline
{\color{blue}\textbf{[Model Prefix]} answer en} {\color{Pos}\textbf{[Task Prefix]} Answer the multiple choice question by only responding the letter of the correct answer.} {\color{red}\textbf{[Taxonomy Prefix]} Prioritize counting objects and quantifying elements over other analysis.} {\color{black}\textbf{[Question]} Answer the multiple choice question by only responding the letter of the correct answer. How many beds are in the image? Select from the following choices.\newline(A) 0\newline(B) 2\newline(C) 1\newline(D) 3\newline(E) 4}
\end{tcolorbox}
\tcbset{reset}
\caption{Example prompt for the CV-Bench Count dataset.}
\label{fig:cvbench_count_example}
\end{figure}
\begin{figure}[h]
\centering
\tcbset{colback=bboxbg}
\begin{tcolorbox}[
    enhanced,
    breakable,
    title=\textsc{What'sUp-A}
    ]
\ttfamily
Image: 
\includegraphics[width=0.5\linewidth]{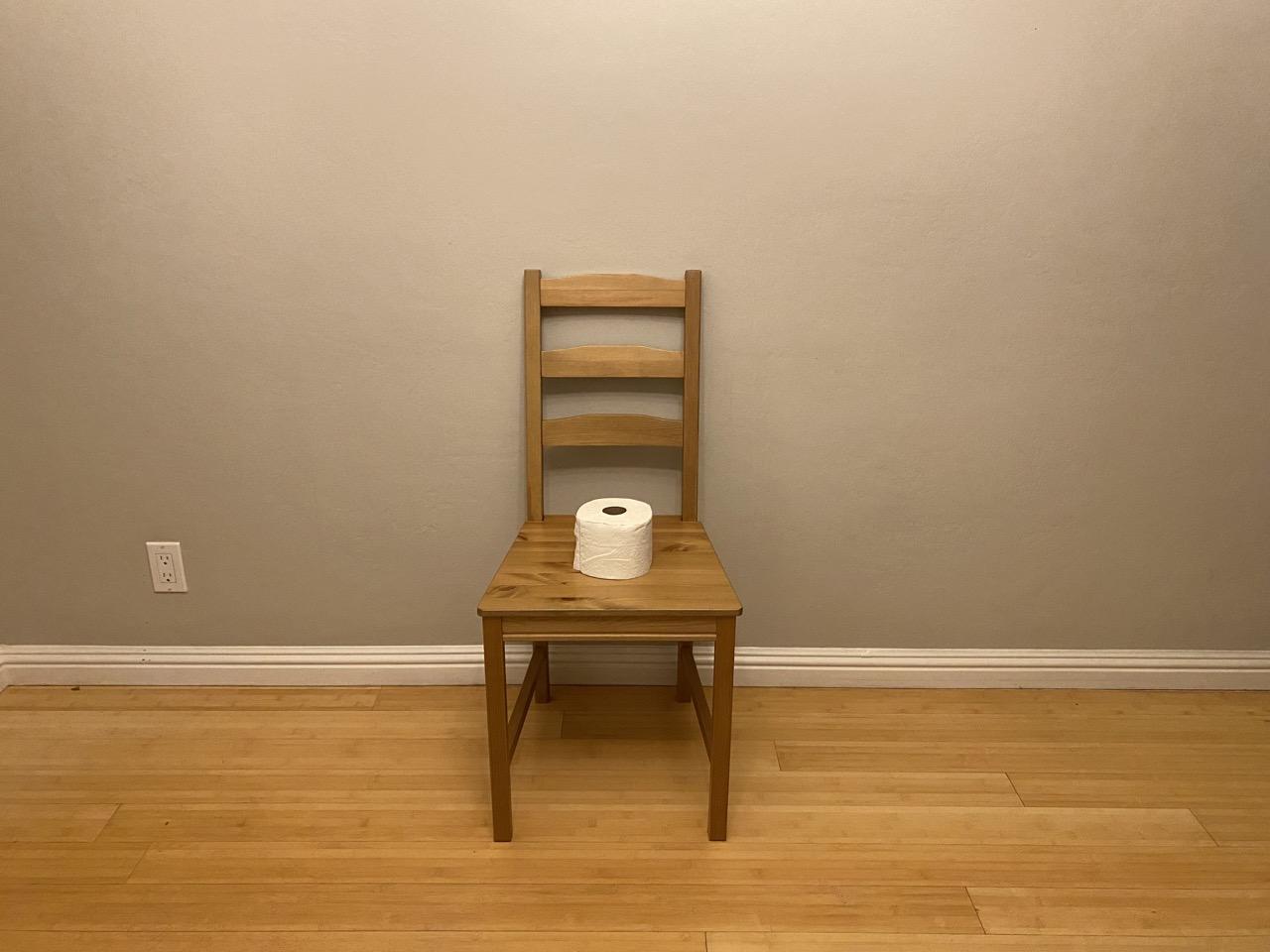}
\newline
{\color{blue}\textbf{[Model Prefix]} answer en} {\color{Pos}\textbf{[Task Prefix]} Answer the multiple choice question by only responding the letter of the correct answer.} {\color{red}\textbf{[Taxonomy Prefix]} Emphasize objects' positions relative to each other.}{\color{black}\textbf{[Question]} Please select the correct caption for the image:\newline(A) A toilet roll under a chair\newline(B) A toilet roll to the left of a chair\newline(C) A toilet roll to the right of a chair\newline(D) A toilet roll on a chair}
\end{tcolorbox}
\caption{Example prompt for the What'sUp-A dataset.}
\label{fig:whatsup_a_example}
\end{figure}
\begin{figure}[ht]
\centering
\tcbset{colback=bboxbg}
\begin{tcolorbox}[title=\textsc{What'sUp-B}]
\ttfamily
Image: 
\includegraphics[width=0.5\linewidth]{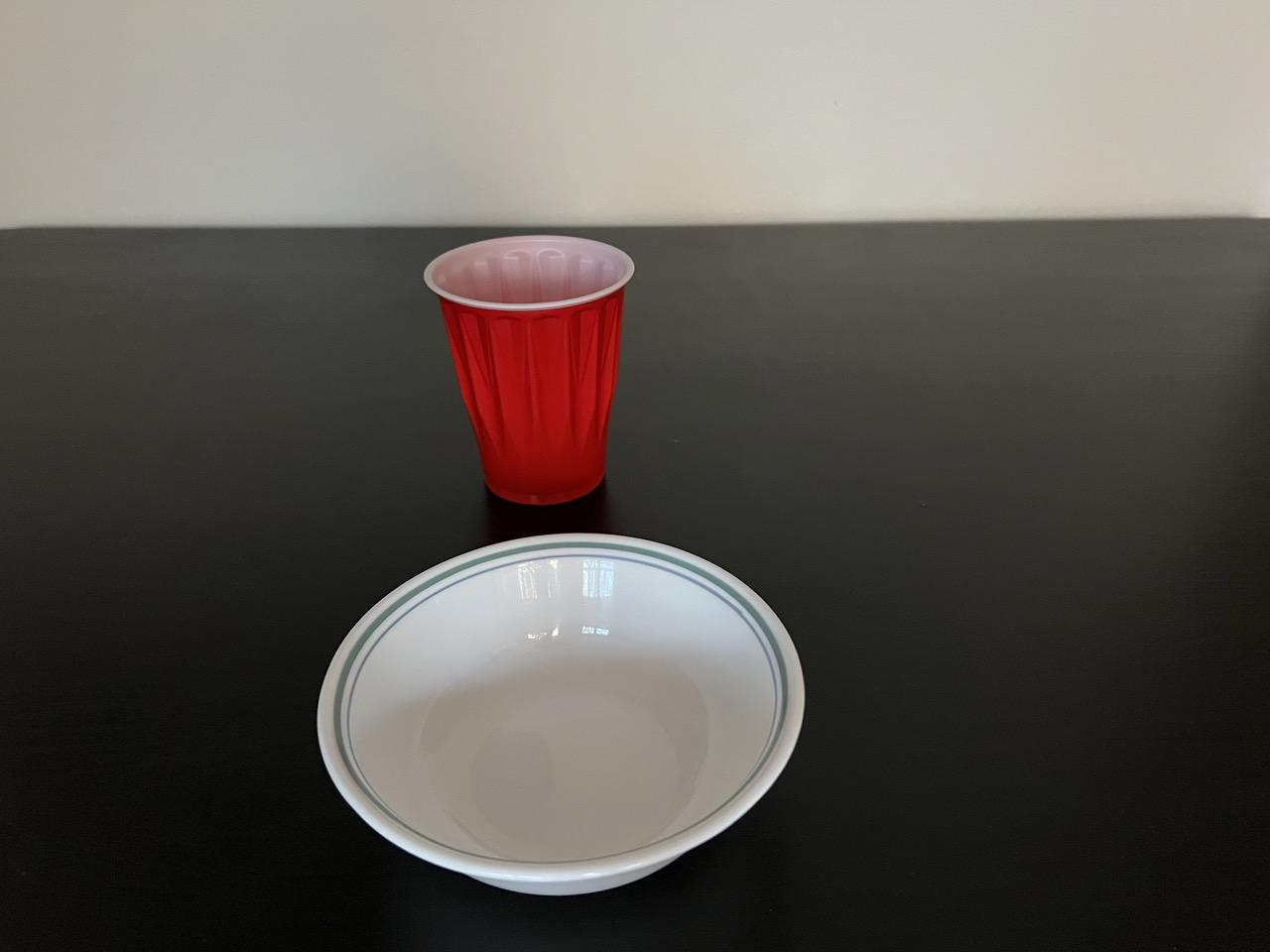}
\newline
{\color{blue}\textbf{[Model Prefix]} answer en} {\color{Pos}\textbf{[Task Prefix]} Answer the multiple choice question by only responding the letter of the correct answer.} {\color{red}\textbf{[Taxonomy Prefix]} Emphasize objects' positions relative to each other.} {\color{black}\textbf{[Question]} Answer the multiple choice question by only responding the letter of the correct answer. Please select the correct caption for the image:\newline(A) A bowl behind a cup\newline(B) A bowl to the left of a cup\newline(C) A bowl to the right of a cup\newline(D) A bowl in front of a cup}
\end{tcolorbox}
\tcbset{reset}
\caption{Example prompt for the What'sUp-B dataset.}
\label{fig:whatsup_b_example}
\end{figure}
\begin{figure}[ht]
\centering
\tcbset{colback=bboxbg}
\begin{tcolorbox}[title=\textsc{BLINK Object Localization}]
\ttfamily
Image: 
\includegraphics[width=0.5\linewidth]{figures/dataset_examples/blink38.jpg}
\newline
{\color{blue}\textbf{[Model Prefix]} answer en} {\color{Pos}\textbf{[Task Prefix]} Answer the multiple choice question by only responding the letter of the correct answer.} {\color{red}\textbf{[Taxonomy Prefix]} Emphasize objects' positions relative to each other.} {\color{black}\textbf{[Question]} A bounding box is an annotated rectangle surrounding an object. The edges of bounding boxes should touch the outermost pixels of the object that is being labeled. Given the two bounding boxes on the image, labeled by A and B, which bounding box more accurately localizes and encloses the teddy bear? Select from the following options.\newline(A) Box A\newline(B) Box B}
\end{tcolorbox}
\tcbset{reset}
\caption{Example prompt for the BLINK Object Localization dataset.}
\label{fig:blink_example}
\end{figure}
\begin{figure}[ht]
\centering
\tcbset{colback=bboxbg}
\begin{tcolorbox}[title=\textsc{CLEVR}]
\ttfamily
Image: 
\includegraphics[width=0.5\linewidth]{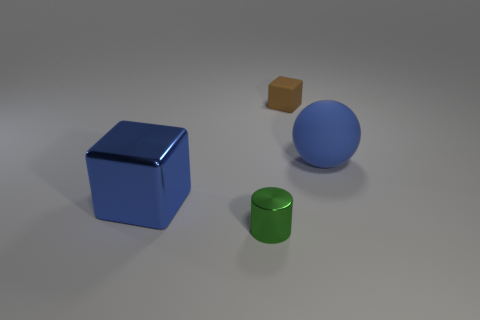}
\newline
{\color{blue}\textbf{[Model Prefix]} answer en} {\color{Pos}\textbf{[Task Prefix]} Answer the question by only responding the number.} {\color{red}\textbf{[Taxonomy Prefix]} Prioritize counting objects and quantifying elements over other analysis.} {\color{black}\textbf{[Question]} How many different items are there in the image?}
\end{tcolorbox}
\tcbset{reset}
\caption{Example prompt for the CLEVR dataset.}
\label{fig:clevr_normal_example}
\end{figure}
\begin{figure}[ht]
\centering
\tcbset{colback=bboxbg}
\begin{tcolorbox}[title=\textsc{Super-CLEVR }]
\ttfamily
Image: 
\includegraphics[width=0.5\linewidth]{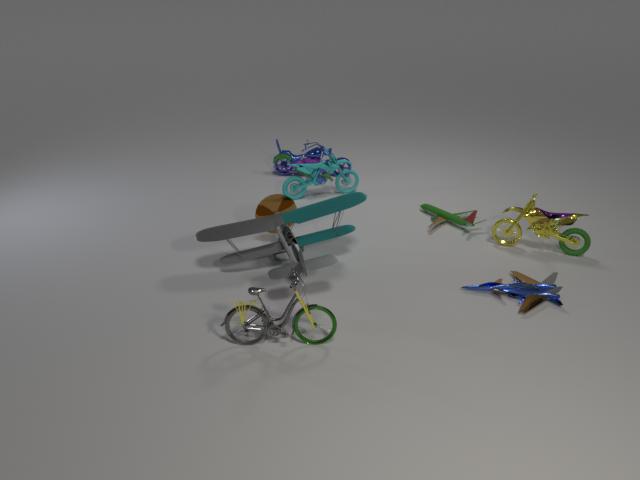}
\newline
{\color{blue}\textbf{[Model Prefix]} answer en} {\color{Pos}\textbf{[Task Prefix]} Answer the question by only responding the number.} {\color{red}\textbf{[Taxonomy Prefix]} Prioritize counting objects and quantifying elements over other analysis.} {\color{black}\textbf{[Question]} How many different items are there in the image?}
\end{tcolorbox}
\tcbset{reset}
\caption{Example prompt for the Super-CLEVR dataset.}
\label{fig:clevr_challenging_example}
\end{figure}
\clearpage
\section{Statistical Significance}
\label{app:statistical_significance}
In this section, we assessed the statistical significance of steering improvements on the What'sUp-A and CLEVR datasets, where we observe the most improvements of steering methods on spatial relation and counting tasks. We used a non-parametric bootstrap approach (10,000 samples) to derive 95\% confidence intervals and computed $p$-values using McNemar's test. Results are presented in Table~\ref{tab:statistical_significance_whatsup} and Table~\ref{tab:statistical_significance_clevr}.
\begin{table}[h]
\centering
\NiceMatrixOptions{cell-space-top-limit=1pt,cell-space-bottom-limit=1pt}
\begin{adjustbox}{width=\textwidth}
\begin{NiceTabular}{l|l|cccc}[code-before = {\rowcolors{3}{gray!12}{white}}]
\toprule
\multirow{2}{*}{\centering\textsc{\textbf{\makecell[l]{Intervention\\Method}}}} & 
\multirow{2}{*}{\centering\textsc{\textbf{Model}}} & 
\multicolumn{4}{c}{\textsc{\textbf{Statistical Significance}}} \\
& & \textsc{Improvement} & \textsc{95\% CI} & \textsc{p-value} & \textsc{Significant?} \\
\midrule
\multirow{3}{*}{\texttt{Prompt}} & 
\texttt{PaliGemma2-3B} & 
{\textcolor{Neg}{-3.4\%}} & 
{[-5.6\%, -1.4\%]} &
{0.00418} & 
{\textcolor{Neg}{No}} \\
& \texttt{PaliGemma2-10B} & 
{\textcolor{Pos}{+2.5\%}} & 
{[-0.3\%, 5.3\%]} &
{0.08326} & 
{\textcolor{Neg}{No}} \\
& \texttt{Idefics3-8B-Llama3} & 
{\textcolor{Pos}{+1.4\%}} & 
{[-0.3\%, 3.4\%]} &
{0.22656} & 
{\textcolor{Neg}{No}} \\
\midrule
\multirow{3}{*}{\texttt{SAE}} & 
\texttt{PaliGemma2-3B} & 
{\textcolor{Pos}{+9.2\%}} & 
{[5.9\%, 12.8\%]} &
{<0.0001} & 
{\textcolor{Pos}{Yes}} \\
& \texttt{PaliGemma2-10B} & 
{\textcolor{Pos}{+11.7\%}} & 
{[8.4\%, 15.4\%]} &
{<0.0001} & 
{\textcolor{Pos}{Yes}} \\
& \texttt{Idefics3-8B-Llama3} & 
{\textcolor{Pos}{+2.0\%}} & 
{[-0.3\%, 4.2\%]} &
{0.14346} & 
{\textcolor{Neg}{No}} \\
\midrule
\multirow{3}{*}{\texttt{Probe}} & 
\texttt{PaliGemma2-3B} & 
{\textcolor{Pos}{+15.9\%}} & 
{[11.7\%, 20.1\%]} &
{<0.0001} & 
{\textcolor{Pos}{Yes}} \\
& \texttt{PaliGemma2-10B} & 
{\textcolor{Pos}{+3.1\%}} & 
{[1.1\%, 5.0\%]} &
{0.00342} & 
{\textcolor{Pos}{Yes}} \\
& \texttt{Idefics3-8B-Llama3} & 
{\textcolor{gray}{+0.0\%}} & 
{[-2.5\%, 2.5\%]} &
{1.00000} & 
{\textcolor{Neg}{No}} \\
\midrule
\multirow{3}{*}{\texttt{MeanShift}} & 
\texttt{PaliGemma2-3B} & 
{\textcolor{Pos}{+12.8\%}} & 
{[8.9\%, 17.0\%]} &
{<0.0001} & 
{\textcolor{Pos}{Yes}} \\
& \texttt{PaliGemma2-10B} & 
{\textcolor{Pos}{+6.4\%}} & 
{[3.6\%, 9.5\%]} &
{<0.0001} & 
{\textcolor{Pos}{Yes}} \\
& \texttt{Idefics3-8B-Llama3} & 
{\textcolor{Neg}{-0.3\%}} & 
{[-4.7\%, 4.2\%]} &
{0.90418} & 
{\textcolor{Neg}{No}} \\
\bottomrule
\end{NiceTabular}
\end{adjustbox}
\vspace{1mm}
\caption{Statistical significance of improvements on the What'sUp-A dataset using bootstrap analysis (10,000 samples) and McNemar's test.}
\label{tab:statistical_significance_whatsup}
\end{table}

\begin{table}[h]
\centering
\NiceMatrixOptions{cell-space-top-limit=1pt,cell-space-bottom-limit=1pt}
\begin{adjustbox}{width=\textwidth}
\begin{NiceTabular}{l|l|cccc}[code-before = {\rowcolors{3}{gray!12}{white}}]
\toprule
\multirow{2}{*}{\centering\textsc{\textbf{\makecell[l]{Intervention\\Method}}}} & 
\multirow{2}{*}{\centering\textsc{\textbf{Model}}} & 
\multicolumn{4}{c}{\textsc{\textbf{Statistical Significance}}} \\
& & \textsc{Improvement} & \textsc{95\% CI} & \textsc{p-value} & \textsc{Significant?} \\
\midrule
\multirow{3}{*}{\texttt{Prompt}} & 
\texttt{PaliGemma2-3B} & 
{\textcolor{Pos}{+2.7\%}} & 
{[0.7\%, 4.7\%]} &
{0.01431} & 
{\textcolor{Pos}{Yes}} \\
& \texttt{PaliGemma2-10B} & 
{\textcolor{Pos}{+2.0\%}} & 
{[-0.2\%, 4.2\%]} &
{0.08326} & 
{\textcolor{Neg}{No}} \\
& \texttt{Idefics3-8B-Llama3} & 
{\textcolor{Pos}{+5.3\%}} & 
{[2.2\%, 8.4\%]} &
{0.00109} & 
{\textcolor{Pos}{Yes}} \\
\midrule
\multirow{3}{*}{\texttt{SAE}} & 
\texttt{PaliGemma2-3B} & 
{\textcolor{Pos}{+18.2\%}} & 
{[14.4\%, 22.0\%]} &
{<0.0001} & 
{\textcolor{Pos}{Yes}} \\
& \texttt{PaliGemma2-10B} & 
{\textcolor{Pos}{+4.2\%}} & 
{[1.6\%, 6.9\%]} &
{0.00235} & 
{\textcolor{Pos}{Yes}} \\
& \texttt{Idefics3-8B-Llama3} & 
{\textcolor{Pos}{+28.2\%}} & 
{[24.0\%, 32.4\%]} &
{<0.0001} & 
{\textcolor{Pos}{Yes}} \\
\midrule
\multirow{3}{*}{\texttt{Probe}} & 
\texttt{PaliGemma2-3B} & 
{\textcolor{Pos}{+4.0\%}} & 
{[1.6\%, 6.7\%]} &
{0.00202} & 
{\textcolor{Pos}{Yes}} \\
& \texttt{PaliGemma2-10B} & 
{\textcolor{Pos}{+0.9\%}} & 
{[-0.7\%, 2.4\%]} &
{0.38770} & 
{\textcolor{Neg}{No}} \\
& \texttt{Idefics3-8B-Llama3} & 
{\textcolor{Pos}{+24.7\%}} & 
{[20.7\%, 28.7\%]} &
{<0.0001} & 
{\textcolor{Pos}{Yes}} \\
\midrule
\multirow{3}{*}{\texttt{MeanShift}} & 
\texttt{PaliGemma2-3B} & 
{\textcolor{Pos}{+14.7\%}} & 
{[11.1\%, 18.4\%]} &
{<0.0001} & 
{\textcolor{Pos}{Yes}} \\
& \texttt{PaliGemma2-10B} & 
{\textcolor{Pos}{+12.0\%}} & 
{[8.7\%, 15.6\%]} &
{<0.0001} & 
{\textcolor{Pos}{Yes}} \\
& \texttt{Idefics3-8B-Llama3} & 
{\textcolor{Pos}{+34.2\%}} & 
{[29.1\%, 39.1\%]} &
{<0.0001} & 
{\textcolor{Pos}{Yes}} \\
\bottomrule
\end{NiceTabular}
\end{adjustbox}
\vspace{1mm}
\caption{Statistical significance of improvements on the CLEVR dataset using bootstrap analysis (10,000 samples) and McNemar's test.}
\label{tab:statistical_significance_clevr}
\end{table}
MeanShift demonstrates strong statistical reliability with significant improvements in 5/6 test cases across both datasets (the lower bound of the confidence interval is greater than 0), achieving substantial gains of up to +34.2\% on CLEVR. SAE shows similar effectiveness with 5/6 significant improvements and particularly strong performance on CLEVR (+28.2\% for Idefics3). Both methods consistently outperform prompting, which shows significance in only 2/6 cases with smaller effect sizes. These results confirm MeanShift and SAE as robust, effective techniques for enhancing visual understanding in MLLMs with high statistical confidence.
\clearpage

\section{Computer Resources} \label{app:compute}
All the experiments discussed in this paper can be done with \textbf{only one} NVIDIA A6000. For faster experiments, we use up to 8 NVIDIA A6000 to run experiments in parallel for various tasks and models.

\section{Broader Impact} \label{app:broader_impact}

The techniques described in this paper are meant to improve the performance of multimodal large language models (MLLMs). Such techniques are capable of being misused to the extent that MLLMs themselves can be misused. We stress that MLLMs and similarly our techniques do not have provable guarantees on the quality or safety of their outputs. 

\section{Licenses} \label{app:licenses}
We list the licenses involved in this work as follows, 
\begin{itemize}
    \item \texttt{PaliGemma2} models and their backbone LLMs \texttt{Gemma2} are under license of \textit{Gemma Terms of Use} \url{https://ai.google.dev/gemma/terms}. 
    \item \texttt{Idefics3-Llama-8B} model is under the license of Apache license 2.0.  Its language backbone, \texttt{Llama-3.1-8B} model, is under the license of  \textit{Llama 3.1 Community License Agreement}. 
    \item \texttt{GemmaScope} pre-trained SAEs are under the license of \textit{Creative Commons Attribution 4.0.} 
    \item \texttt{LllamaScope} pre-trained SAEs are under the license of \textit{Apache License 2.0. }
    \item \texttt{CV-Bench} is under the license of \textit{Apache License 2.0}.
    \item \texttt{What'sUp} datasets are under the license of \emph{MIT License}. 
    \item \texttt{BLINK} dataset is under the license of the \textit{Apache License 2.0}.  
    \item \texttt{CLEVR} and \texttt{Super-CLEVER} datasets are under the \textit{Creative Commons CC BY 4.0} license and the \textit{BSD License}, respectively. 
    \item Our usage of OpenAI's models for prompting is under the license of OpenAI's Terms of Service. 
\end{itemize}

% \clearpage 
% \input{neurips/checklist}

\end{document}